\PassOptionsToPackage{table}{xcolor}
\documentclass{article} 
\usepackage{iclr2027_conference,times}

\usepackage{amsmath,amsfonts,bm}

\def\eqref#1{equation~\ref{#1}}

\def\1{\bm{1}}

\DeclareMathAlphabet{\mathsfit}{\encodingdefault}{\sfdefault}{m}{sl}
\SetMathAlphabet{\mathsfit}{bold}{\encodingdefault}{\sfdefault}{bx}{n}

\usepackage{hyperref}
\usepackage{url}
\usepackage{booktabs}
\usepackage{enumitem}
\usepackage{graphicx}
\usepackage{caption}
\usepackage{xcolor}
\usepackage{amssymb}
\usepackage{listings}
\usepackage{tikz}
\definecolor{promptbg}{RGB}{248,250,252}
\definecolor{promptrule}{RGB}{190,201,214}
\definecolor{promptsystem}{RGB}{0,91,150}
\definecolor{promptuser}{RGB}{91,75,138}
\definecolor{promptoutput}{RGB}{0,120,102}
\definecolor{promptslot}{RGB}{157,78,0}
\lstdefinestyle{synthesisprompt}{
    basicstyle=\ttfamily\fontsize{7.35pt}{9.0pt}\selectfont,
    breaklines=true,
    breakatwhitespace=true,
    columns=fullflexible,
    keepspaces=true,
    showstringspaces=false,
    sensitive=true,
    morekeywords=[1]{SYSTEM},
    morekeywords=[2]{USER},
    morekeywords=[3]{OUTPUT},
    keywordstyle=[1]\color{promptsystem}\bfseries,
    keywordstyle=[2]\color{promptuser}\bfseries,
    keywordstyle=[3]\color{promptoutput}\bfseries,
    moredelim=**[s][\color{promptslot}]{\{} {\}},
    frame=single,
    framerule=0.45pt,
    rulecolor=\color{promptrule},
    backgroundcolor=\color{promptbg},
    framesep=5pt,
    xleftmargin=1pt,
    xrightmargin=1pt,
    aboveskip=4pt,
    belowskip=7pt
}

\title{Think in 360°: Benchmarking and 
Bootstrapping Panoramic Video Reasoning}

\author{Lidong Lu$^{1}$, Lidong Zou$^{1}$, Xinyi Mao$^{1}$, Jiaxin Lv$^{1}$, 
Zhenquan Li$^{1}$, Liangrui Dong$^{1}$, \\ \textbf{Tong Lu$^{1}$, Guo Chen$^{2}$} \\[3pt]
\normalfont $^{1}$Nanjing University \qquad $^{2}$NVIDIA \\[4pt]
\makebox[\dimexpr\textwidth-2\tabcolsep\relax][c]{\small  \url{https://think-in-360.github.io}}}

\definecolor{ExampleBlue}{HTML}{0072B2}
\definecolor{ExampleTeal}{HTML}{008A70}
\definecolor{ExampleGold}{HTML}{D58B00}
\definecolor{ExampleInk}{HTML}{20242B}
\definecolor{ExampleMuted}{HTML}{626A73}
\definecolor{ExampleRule}{HTML}{CAD4DC}
\definecolor{ExampleQuestionWash}{HTML}{F3F8FB}
\definecolor{ExampleFrameWash}{HTML}{FBFCFD}
\definecolor{ExampleReasonWash}{HTML}{F0F7FA}
\definecolor{ExampleAnswerWash}{HTML}{EDF8F4}
\definecolor{ExampleError}{HTML}{D55E00}
\definecolor{ExampleErrorWash}{HTML}{FDF3EE}
\definecolor{ExampleOptionWash}{HTML}{E5EFF5}
\definecolor{ExampleTimeWash}{HTML}{EAF2F7}

\newcommand{\ExampleLabel}[2]{%
  \noindent\textcolor{#1}{\rule{3pt}{1.15em}}\hspace{5pt}%
  {\sffamily\bfseries\small\color{ExampleInk}#2}\par%
  \vspace{2pt}{\color{ExampleRule}\hrule height 0.35pt}\vspace{3pt}%
}

\newcommand{\ExamplePanelAccent}[4]{%
  \par\noindent
  \begingroup
  \setlength{\fboxsep}{5pt}%
  \setlength{\fboxrule}{0.55pt}%
  \fcolorbox{ExampleRule}{#1}{%
    \parbox{\dimexpr\linewidth-2\fboxsep-2\fboxrule\relax}{%
      \ExampleLabel{#2}{#3}#4}}%
  \endgroup\par
}

\newcommand{\ExamplePanel}[3][white]{%
  \ExamplePanelAccent{#1}{ExampleBlue}{#2}{#3}%
}

\newcommand{\ExampleEvidencePanel}[1]{%
  \ExamplePanelAccent{ExampleFrameWash}{ExampleGold}{Selected frames}{#1}%
}

\newcommand{\ExampleTimeline}[2][19bp]{%
  \centering
  \includegraphics[width=\linewidth,trim=0 0 0 #1,clip]{#2}\par
}

\newcommand{\ExampleOptionLine}[2]{%
  \par\noindent\hangindent=1.7em\hangafter=1%
  {\setlength{\fboxsep}{1.2pt}%
   \colorbox{ExampleOptionWash}{\sffamily\bfseries\scriptsize\color{ExampleBlue}#1}}%
  \hspace{0.35em}#2\par%
}

\newcommand{\ExampleOptions}[4]{%
  \begin{minipage}[t]{0.485\linewidth}\footnotesize
    \ExampleOptionLine{A}{#1}\vspace{3pt}%
    \ExampleOptionLine{C}{#3}%
  \end{minipage}\hfill
  \begin{minipage}[t]{0.485\linewidth}\footnotesize
    \ExampleOptionLine{B}{#2}\vspace{3pt}%
    \ExampleOptionLine{D}{#4}%
  \end{minipage}%
}

\newenvironment{QualitativeExample}{%
  \par\noindent\hspace*{0.0125\textwidth}%
  \begin{minipage}{0.975\textwidth}
  \color{ExampleInk}%
}{%
  \end{minipage}\par%
}

\newcommand{\ExampleReasoning}[1]{%
  \ExamplePanelAccent{ExampleReasonWash}{ExampleBlue}{PanoVideoReasoner reasoning}{%
    {\footnotesize\color{ExampleInk}#1}\par}%
}

\newcommand{\ExampleComparison}[2]{%
  \vspace{4pt}%
  \ExamplePanelAccent{white}{ExampleTeal}{Model outputs}{%
    \noindent
    \begin{minipage}[t]{0.485\linewidth}
      \setlength{\fboxsep}{5pt}%
      \colorbox{ExampleAnswerWash}{%
        \parbox{\dimexpr\linewidth-2\fboxsep\relax}{%
          {\sffamily\bfseries\scriptsize\color{ExampleTeal}PanoVideoReasoner \checkmark}\par
          \vspace{2pt}{\footnotesize\color{ExampleInk}#1}\par}}
    \end{minipage}\hfill
    \begin{minipage}[t]{0.485\linewidth}
      \setlength{\fboxsep}{5pt}%
      \colorbox{ExampleErrorWash}{%
        \parbox{\dimexpr\linewidth-2\fboxsep\relax}{%
          {\sffamily\bfseries\scriptsize\color{ExampleError}Qwen3.8-27B $\times$}\par
          \vspace{2pt}{\footnotesize\color{ExampleInk}#2}\par}}
    \end{minipage}\par}%
}

\newcommand{\ExampleCaption}[2]{%
  \vspace{5pt}{\color{ExampleBlue}\hrule height 0.6pt}\vspace{2pt}%
  \begingroup
  \captionsetup{font=small,labelfont=bf,skip=0pt}%
  \captionof{figure}{#1}\label{#2}%
  \endgroup
}

\iclrfinalcopy
\begin{document}

\maketitle
\fancyhead{}
\renewcommand{\headrulewidth}{0pt}

\begin{abstract}
A panoramic video captures all directions around the camera, but relevant evidence can appear in different directions and at different times. Understanding it requires models to connect observations across the viewing sphere and time. We introduce PanoVideoBench, the first manually annotated benchmark for comprehensive panoramic video understanding. Its 1,000 questions span five categories and 15 question types, with diagnostic annotations for target search and panoramic capabilities. Evaluation of 11 proprietary, open-source, and panoramic-specialized models identifies cross-view tracking and sphere-time evidence integration as their weakest capabilities. To strengthen these capabilities, we construct PanoVideoCoT, the first reasoning dataset grounded in real panoramic video, through sphere-time-grounded synthesis that converts verified panoramic observations and successful agentic explorations into 13,465 QA--CoT pairs. Starting from a general video MLLM, PanoVideoReasoner learns from this corpus through Panoramic CoT SFT and Evidence-Privileged On-Policy Self-Distillation, a panoramic instantiation of Vision-OPD's OPSD in which the teacher uses verified perspective views to guide student-generated trajectories. It reaches 55.7\% on PanoVideoBench, improving the backbone by 12.5 points and the previous SOTA by 2.6. The gains concentrate on the same spatial, cross-view, and temporal capabilities exposed by the benchmark, showing that grounded sphere-time supervision can equip a general video MLLM with panoramic reasoning.

\end{abstract}

\begin{figure*}[h]
    \centering
    \includegraphics[width=\textwidth]{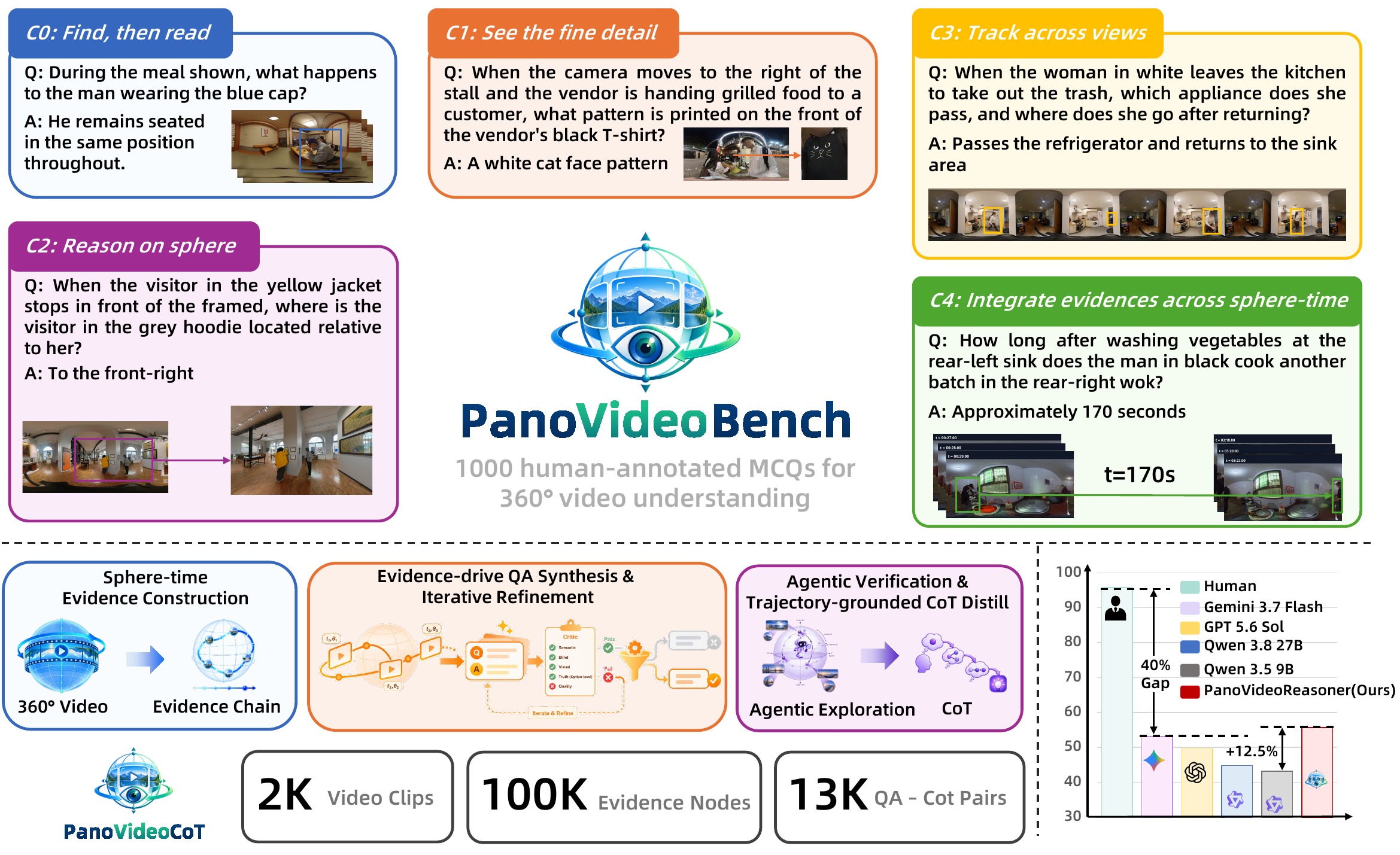}
    \caption{
    \textbf{A 360$^\circ$ video shows the whole scene, but the evidence can still be hard to find.}
    PanoVideoBench evaluates how models find and connect evidence across direction and time. PanoVideoCoT turns verified sphere-time evidence and solution trajectories into reasoning supervision, improving Qwen3.5-9B by 12.5 points and establishing a new SOTA.
    }
    \label{fig:teaser}
\end{figure*}

\section{Introduction}

Panoramic video preserves an entire environment rather than a camera-selected view, supporting immersive media and applications that must observe beyond a single field of view, including telepresence, robotics, and scene monitoring. This broader record changes the understanding problem. Perspective video begins with a choice made by the camera: where to look. Its limited view usually keeps the main action in frame and narrows the model's search. A panoramic image preserves the surrounding scene at one moment. In panoramic video, relevant evidence can emerge anywhere as the scene evolves. The model must therefore find, retain, and connect observations across direction and time.

Video benchmarks now cover compositional QA and increasingly long temporal contexts~\cite{xiao2021nextqa,grundemclaughlin2021agqa,li2023mvbench,chen2024cgbench,fu2024videomme,wang2024lvbench,wu2024longvideobench,ataallah2024infinibench}. More targeted evaluations test whether models actually use temporal order, cross-segment evidence, and explicit reasoning chains~\cite{cores2024tvbench,cai2024temporalbench,shangguan2024tomato,chen2024rextime,cheng2025vstar,qi2025vcrbench}. However, their perspective videos already direct the model toward the recorded action. Panoramic datasets and models study 360$^\circ$ perception, spherical relations, view selection, and multi-step spatial reasoning~\cite{chou2020vqa360,yun2021panoavqa,chen2024360x,dongfang2025osr,zhang2025360r1,zhou2025dense360,he2026omnicot,wang2026panoworld}, but most focus on images or localized events. Continuous evidence search across direction and time remains largely unexplored.

To evaluate this ability, we introduce PanoVideoBench, the first manually annotated benchmark for comprehensive panoramic video understanding. Its 1,000 four-choice questions span five categories and 15 question types. Two diagnostic axes separate how the evidence is found from the panoramic operations needed to answer. Across eleven proprietary, open-source, and panoramic models, performance drops most when evidence must be retained across views or time.

The weaknesses exposed by PanoVideoBench point to a way to improve general video MLLMs: train them to reason over evidence distributed across views and time. We therefore construct PanoVideoCoT through sphere-time-grounded CoT synthesis. The pipeline organizes video into an entity-linked sphere-time evidence memory, builds questions and plausible options from supporting relations, and refines them with specialized visual critics. A visual agent solves each question through blind exploration, followed when necessary by evidence-guided exploration. Successful trajectories are reorganized into concise CoTs, yielding 13,465 QA--CoT pairs from 2,063 videos and approximately 100K evidence nodes.

Using PanoVideoCoT, we train PanoVideoReasoner in two stages: Panoramic CoT SFT and Evidence-Privileged On-Policy Self-Distillation (EP-OPSD), a panoramic instantiation of Vision-OPD's OPSD. SFT learns from the synthesized CoTs, while EP-OPSD retains the original distillation objective and uses verified evidence views as the teacher's privileged context. On the same QA pairs, CoT supervision adds 6.2 points over answer-only SFT, and EP-OPSD adds another 3.9 points over CoT SFT. PanoVideoReasoner reaches 55.7\% on PanoVideoBench, improving its Qwen3.5-9B backbone by 12.5 points and the previous SOTA by 2.6.

\section{Related Work}

\paragraph{Panoramic visual understanding benchmarks.} Panoramic vision spans perception, multimodal understanding, and embodied spatial intelligence~\cite{zheng2025panorama,lin2025panoramicsurvey}. Its benchmarks have progressed from question answering over surrounding images~\cite{chou2020vqa360} to fine-grained perception~\cite{zhou2025dense360,yang2025odibench,tran2026360bench}, spherical and 3D spatial relations~\cite{dongfang2025osr,zhang2025360r1,lin2026panoenv}, correlated panoramas~\cite{zhang2025cfpano}, and multi-step spatial reasoning~\cite{he2026omnicot}. Panoramic video benchmarks remain specialized: Pano-AVQA evaluates spherical audio-visual relations~\cite{yun2021panoavqa}, while ST-OmniQA focuses on recognizing, localizing, and tracking sound sources~\cite{zeng2026stomniqa}. As Table~\ref{tab:compact_benchmark_comparison} shows, PanoVideoBench is the first manually annotated benchmark to evaluate panoramic video understanding across appearance, space, action, motion, and time.

\begin{table*}[t]
    \centering
    \caption{\textbf{Comparison with existing panoramic benchmarks.} The table separates input modality, evaluation size, manual annotation, and task coverage. Manual denotes direct human annotation of the evaluation questions and gold answers. I, V, and A denote image, video, and audio; C360I denotes correlated panoramic images, and FOA denotes first-order Ambisonics. APP, SPA, ACT, MOT, TEM, and AV denote appearance and detail perception, spatial and viewpoint reasoning, action and state understanding, motion and identity tracking, temporal event reasoning, and audio-visual reasoning. Checkmarks indicate an explicitly evaluated dimension, and \#Eval reports the number of evaluation questions.}
    \label{tab:compact_benchmark_comparison}
    \small
    \renewcommand{\arraystretch}{1.30}
    \setlength{\tabcolsep}{7pt}
    \resizebox{\textwidth}{!}{%
    \begin{tabular}{lccc|cccccc}
        \toprule
        Benchmark & Input & \#Eval & Manual & APP & SPA & ACT & MOT & TEM & AV \\
        \midrule
        VQA 360$^\circ$~\cite{chou2020vqa360} & 360I & 7.0K & -- & \checkmark & \checkmark & -- & -- & -- & -- \\
        360Bench~\cite{tran2026360bench} & 360I & 1.5K & \checkmark & \checkmark & \checkmark & -- & -- & -- & -- \\
        CFpano~\cite{zhang2025cfpano} & C360I & 1.6K & -- & \checkmark & \checkmark & -- & -- & -- & -- \\
        OmniCoT-B~\cite{he2026omnicot} & 360I & 6.7K & -- & -- & \checkmark & -- & -- & -- & -- \\
        Pano-AVQA~\cite{yun2021panoavqa} & 360V+A & 5.3K & -- & -- & \checkmark & -- & -- & -- & \checkmark \\
        ST-OmniQA~\cite{zeng2026stomniqa} & 360V+FOA & 4.0K & -- & -- & \checkmark & -- & \checkmark & \checkmark & \checkmark \\
        \textbf{PanoVideoBench (Ours)} & 360V & 1.0K & \checkmark & \checkmark & \checkmark & \checkmark & \checkmark & \checkmark & -- \\
        \bottomrule
    \end{tabular}
    }
\end{table*}

\paragraph{Panoramic training data.} General video reasoning datasets construct intermediate supervision through active annotation or core-frame selection~\cite{wang2024videocot,han2024videoespresso}. Panoramic datasets supervise surrounding-scene understanding through region-grounded answers, dense entity descriptions, referring expressions, or spatial questions derived from 3D scenes~\cite{yun2021panoavqa,zhou2025dense360,lin2026panoenv}. Reasoning-oriented datasets build CoTs from panoramic descriptions or serialized 3D scenes, while EAGLE-360 records exploration trajectories for active object search~\cite{zhang2025360r1,he2026omnicot,xu2026eagle360}. ST-OmniQA provides executable traces for sound-source reasoning in synthetic panoramic videos~\cite{zeng2026stomniqa}. PanoVideoCoT instead grounds both questions and CoTs in real panoramic video: questions arise from verified sphere-time relations, and CoTs are reconstructed from successful visual exploration.

\begin{table*}[t]
    \centering
    \caption{\textbf{Comparison with existing panoramic training datasets.} The table compares source scale and modality, whether the visual data are real, the forms of supervision, and how questions and solutions are constructed. \#Vis. and \#Train report visual sources and training instances. VG, TG, E-QG, and ReSolve denote visual grounding, temporal grounding, evidence-grounded question generation, and independent visual re-solving; checkmarks indicate that a property is provided.}
    \label{tab:training_data_comparison}
    \renewcommand{\arraystretch}{1.30}
    \resizebox{\textwidth}{!}{%
    \begin{tabular}{lcc|cc|ccc|cc}
        \toprule
        & &
        & \multicolumn{2}{c|}{Data basis}
        & \multicolumn{3}{c|}{Supervision}
        & \multicolumn{2}{c}{Construction} \\
        \cmidrule(lr){4-5}
        \cmidrule(lr){6-8}
        \cmidrule(lr){9-10}
        Dataset & \#Vis. & \#Train
        & 360V & Real
        & CoT & VG & TG
        & E-QG & ReSolve \\
        \midrule
        Pano-AVQA~\cite{yun2021panoavqa}
        & 4.5K clips & 42.8K QA
        & \checkmark & \checkmark & -- & \checkmark & -- & \checkmark & -- \\
        OmniVQA~\cite{zhang2025360r1}
        & 1.2K images & 4.9K QA
        & -- & \checkmark & \checkmark & -- & -- & -- & -- \\
        Dense360~\cite{zhou2025dense360}
        & 160K images & 5M ent. cap.
        & -- & \checkmark & -- & \checkmark & -- & -- & -- \\
        PanoEnv-QA~\cite{lin2026panoenv}
        & 0.4K images & 10.3K QA
        & -- & -- & -- & \checkmark & -- & \checkmark & -- \\
        OmniCoT-T~\cite{he2026omnicot}
        & 2.8K images & 14.3K QA--CoT
        & -- & -- & \checkmark & \checkmark & -- & \checkmark & -- \\
        EAGLE-360~\cite{xu2026eagle360}
        & 13.0K images & 13.0K traj.
        & -- & \checkmark & \checkmark & \checkmark & -- & \checkmark & -- \\
        ST-OmniQA~\cite{zeng2026stomniqa}
        & 39.6K videos & 396K QA
        & \checkmark & -- & \checkmark & \checkmark & \checkmark & \checkmark & -- \\
        \midrule
        \textbf{PanoVideoCoT (Ours)}
        & 2.1K videos & 13.5K QA-CoT
        & \checkmark & \checkmark & \checkmark & \checkmark & \checkmark & \checkmark & \checkmark \\
        \bottomrule
    \end{tabular}%
    }
\end{table*}

\paragraph{Models for panoramic understanding.} Panoramic MLLMs mainly adapt image models to ERP geometry through spherical representations, panorama-aware position encoding, view projection, or tool-based exploration~\cite{dongfang2025osr,zhang2025360r1,zhou2025dense360,tran2026360bench,wang2026panoworld,he2026omnicot,xu2026eagle360}. These designs improve panoramic perception and spatial reasoning, but largely remain tied to specialized architectures or static-image tasks. General video MLLMs now support longer inputs, denser temporal modeling, and more efficient frame representations~\cite{maaz2023videochatgpt,lin2023videollava,li2023llamavid,ren2023timechat,qian2024momentor,xu2024pllava,wang2024groundedvideollm}. PanoVideoReasoner takes the more general route: PanoVideoCoT supplies panoramic reasoning supervision, and an evidence-privileged adaptation of OPSD transfers it to a general video MLLM~\cite{yuan2026visionopd}.

\section{PanoVideoBench: Comprehensive Evaluation of Panoramic Video Understanding}\label{sec:benchmark}
\subsection{Benchmark Construction}

We construct PanoVideoBench from videos in 360+x~\cite{chen2024360x}. The process has two stages: question annotation in an interactive spherical player and independent review of every item.

\paragraph{Spherical-video annotation.} The interface pairs a draggable spherical player with a synchronized ERP view. Annotators scrub the timeline and rotate the viewing direction to follow events or inspect objects anywhere on the sphere. The ERP preserves global context, the perspective window reveals local detail, and an orientation aid supports questions about direction, facing, and motion. After exploring the video across direction and time, an annotator records a question, its correct answer, and three plausible distractors. Annotators therefore establish both where and when to look before writing an item.

\paragraph{Independent review.} Two additional annotators independently replay the full spherical video and answer each question. Together with the initial label, their judgments form three votes, and the majority choice becomes the gold answer. Reviewers verify that the selected answer is uniquely supported by visible evidence and that the remaining options are plausible but false. Direction-sensitive items receive an additional reference-frame check. Text-only screening removes questions that can be solved without the video. Automatic checks remove malformed or near-duplicate items, verify answer remapping after option shuffling, and balance the correct positions across A--D. Among 1,296 questions that pass initial screening, 983 receive unanimous votes. A final adjudication then retains 17 of the 171 questions that require majority voting, completing the 1,000-question benchmark. Appendix~\ref{app:quality_control} details the full quality-control protocol.

\subsection{Question Taxonomy and Diagnostic Annotations}

\paragraph{Semantic taxonomy.} Panoramic video understanding begins with recognizing what surrounds the camera, then extends to how that content is arranged and how it changes. We organize this space into five categories and 15 question types. Panoramic Appearance and Detail Perception (APP) covers the scene, its entities, and fine visual details. Spherical Spatial and Viewpoint Reasoning (SPA) turns the complete viewing sphere into spatial relations and viewpoint transformations. For dynamic content, Omnidirectional Action and State Understanding (ACT) captures activities and state changes, Cross-View Motion and Identity Tracking (MOT) follows entities across direction and time, and Panoramic Spatio-Temporal Event Reasoning (TEM) connects events through localization, ordering, counting, and aggregation. Appendix~\ref{app:question_taxonomy} gives the full definitions.

\paragraph{Diagnostic annotations.} Each semantic label identifies the visual task posed by the question. The two diagnostic axes then follow the evidence needed to solve it. Direction-Guided Search (S0), Landmark-Guided Search (S1), Event-/Moving-Entity-Guided Search (S2), and Unconstrained Panoramic Search (S3) record how the model is led toward the relevant view. Find, then Read (C0), See the Fine Detail (C1), Reason on the Sphere (C2), Track Across Views (C3), and Integrate Evidence Across Sphere--Time (C4) describe what the model does with the evidence it finds. Each question has one search setting. C0 marks questions resolved by local readout, while C1--C4 can co-occur when several panoramic operations are needed. Together, the three views show whether the model understands the task, reaches the evidence, and connects the necessary observations.

\subsection{Benchmark Statistics}
\begin{figure*}[t]
    \centering
    \includegraphics[width=\textwidth]{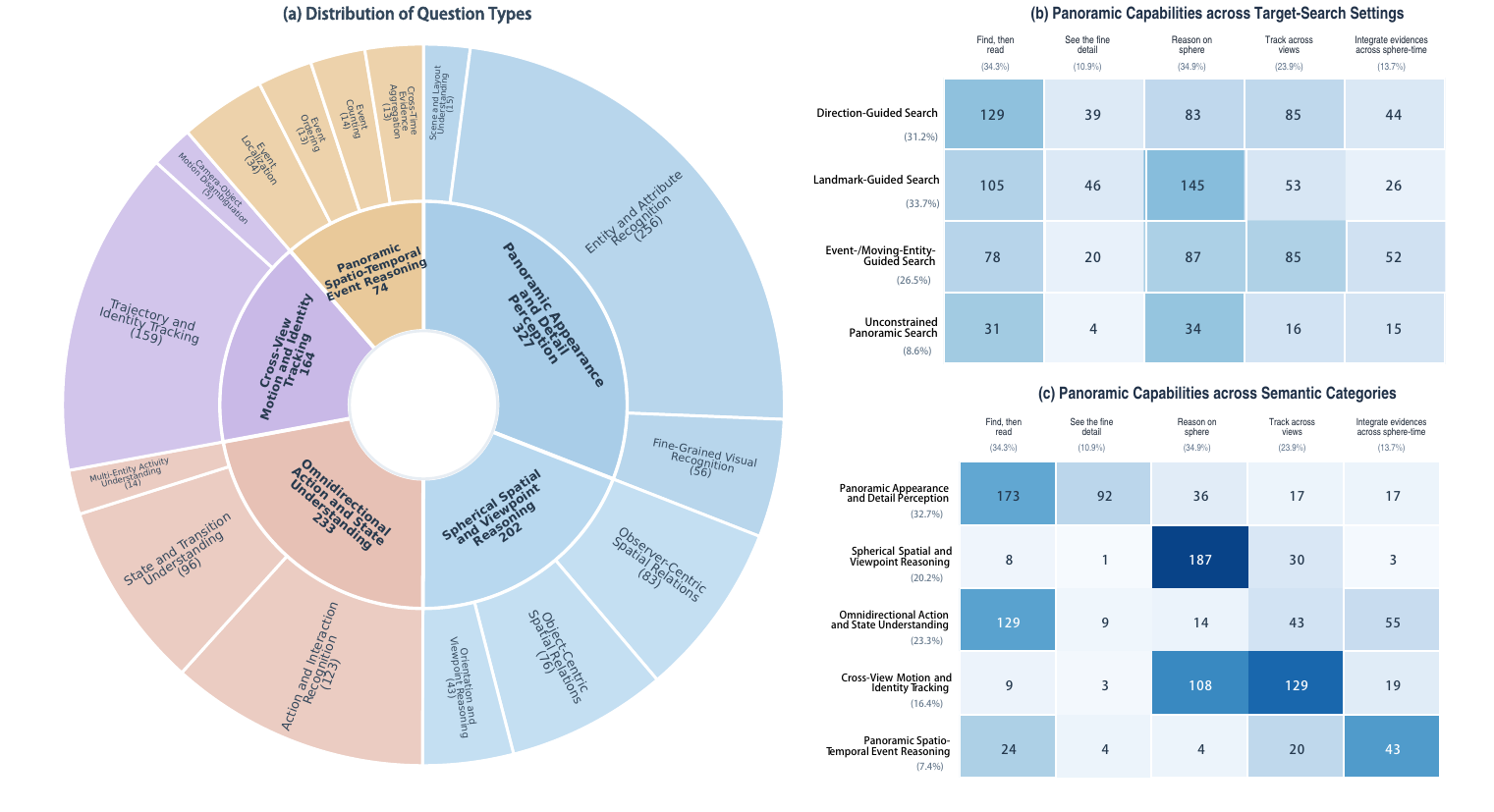}
    \caption{
\textbf{Question composition and diagnostic coverage in PanoVideoBench.}
The sunburst reports the five semantic categories and 15 question types. The heatmaps cross evidence-search settings and semantic categories with five panoramic capabilities. Percentages indicate dataset-wide frequency, and cells give question counts.
}
    \label{fig:benchmark_statistics}
\end{figure*}
PanoVideoBench contains 1,000 questions from 223 videos, spanning 23.06 hours with an average duration of 372 seconds. Appearance and detail form the largest category at 32.7\%, while the three dynamic categories together account for 47.1\%. The benchmark therefore covers both discovering content around the camera and following how it changes across directions and time. Only 31.2\% of questions directly name a viewing direction; the remainder use a landmark or dynamic anchor, or provide no spatial cue, making evidence search part of the task.

The category--capability matrix separates what a question asks from the operations needed to answer it. APP is not synonymous with fine-grained recognition: many questions reduce to local readout after the relevant view is found, while others require resolving small objects or relating evidence across directions and time. SPA has a sharper signature, with 187 of 202 questions requiring sphere reasoning. ACT ranges from local actions to activities whose participants or states must be followed across views. MOT couples sphere reasoning with cross-view tracking because movement direction and identity continuity become inseparable when an entity crosses viewing directions. TEM emphasizes sphere--time integration, but also includes local-readout and tracking cases. The semantic categories thus identify answer targets rather than forming a difficulty ladder.

The search--capability matrix captures a separate bottleneck. A landmark may locate the relevant region without resolving the spatial relations within it, while a moving cue often leads to tracking and sphere--time integration. Likewise, an explicit direction narrows the initial search but does not remove subsequent relational reasoning. Conversely, an unconstrained search can end in simple local readout once the target is found. Broad search need not imply complex inference, nor does a precise cue imply an easy question. The semantic category states what must be answered, the search setting describes how evidence is reached, and the capability labels capture how observations must be connected.

\section{PanoVideoReasoner: Learning Panoramic Video Reasoning}
\subsection{Overview}

A general video MLLM can recognize actions and temporal events, yet panoramic video requires reasoning over observations separated in direction and time. PanoVideoReasoner builds this ability in a general video MLLM through panoramic reasoning supervision. Our synthesis pipeline organizes panoramic observations in a shared sphere-time structure, constructs questions from relations within it, and turns successful agentic explorations into CoTs. This process produces \textbf{PanoVideoCoT}. PanoVideoReasoner first learns complete reasoning trajectories from its grounded CoTs, then improves the trajectories it generates through evidence-guided on-policy supervision.

\subsection{Sphere-Time Grounded CoT Synthesis}

PanoVideoCoT connects three stages through a shared sphere-time grounding. We organize the video into entity-centered evidence across directions and time, turn selected evidence relations into questions, and reconstruct the CoT from an agent's successful exploration of the video.

\paragraph{Sphere-time evidence memory.} We retain the full ERP frame at each sampled timestamp and inspect directional perspective views for entities, attributes, actions, events, and states. A \emph{sphere-time evidence node} represents one entity through multiple observations. Each observation stores its timestamp or interval, yaw, pitch, and supporting view. A VLM associates observations across views and clips using appearance, scene context, and motion continuity, allowing the memory to follow the same entity as its direction and state change. A question may draw on one observation for local perception or connect several observations for spherical, temporal, identity, and state-transition reasoning.

\paragraph{Evidence-driven question construction.} Each question begins with a relation supported by selected observations. This relation fixes the answer, while distractors reuse entities and events from the same video but alter the decisive attribute, direction, time, or identity. Specialized critics test the evidence relation, the visual validity of all four options, language-only shortcuts, and the clarity of the question. Their feedback updates the affected question component before the candidate is reviewed again. We describe the angular extent of its evidence by $\mathrm{AES}=360^\circ-g_{\max}$, where $g_{\max}$ is the largest gap between consecutive evidence directions.

\paragraph{Agentic exploration and CoT construction.} We obtain the CoT from a successful video-grounded solution process. The agent first explores the ERP video from the question and options, choosing when and where to inspect. Questions not solved in this blind stage enter a second stage guided by textual evidence descriptions with their time, yaw, and pitch. Each stage runs at most four rollouts and stops at the first correct answer; the agent is encouraged to inspect the supporting sphere-time region before committing to an option. A trajectory distiller identifies the decisive inspections, revised expectations, and option eliminations, and a writer reorganizes them into a concise CoT. Certification checks that the CoT supports the answer and that its visual statements match the inspected time and direction. Each  sample contains the ERP video, question, options, answer, CoT, and verified evidence with its descriptions, intervals, directions, and perspective views. Of approximately 32K question skeletons, 18K pass critic refinement. Blind exploration solves roughly 8K and evidence-guided exploration recovers another 5K, yielding 13,465 QA--CoT pairs.

\subsection{PanoVideoCoT Statistics}

The synthesis pipeline produces 13,465 QA--CoT pairs from 2,063 panoramic videos in 360-1M~\cite{wallingford2024image}, spanning 168.8 hours. It constructs approximately 100K sphere-time evidence nodes across viewing directions and timestamps.

\begin{figure*}[t]
    \centering
    \includegraphics[width=\textwidth]{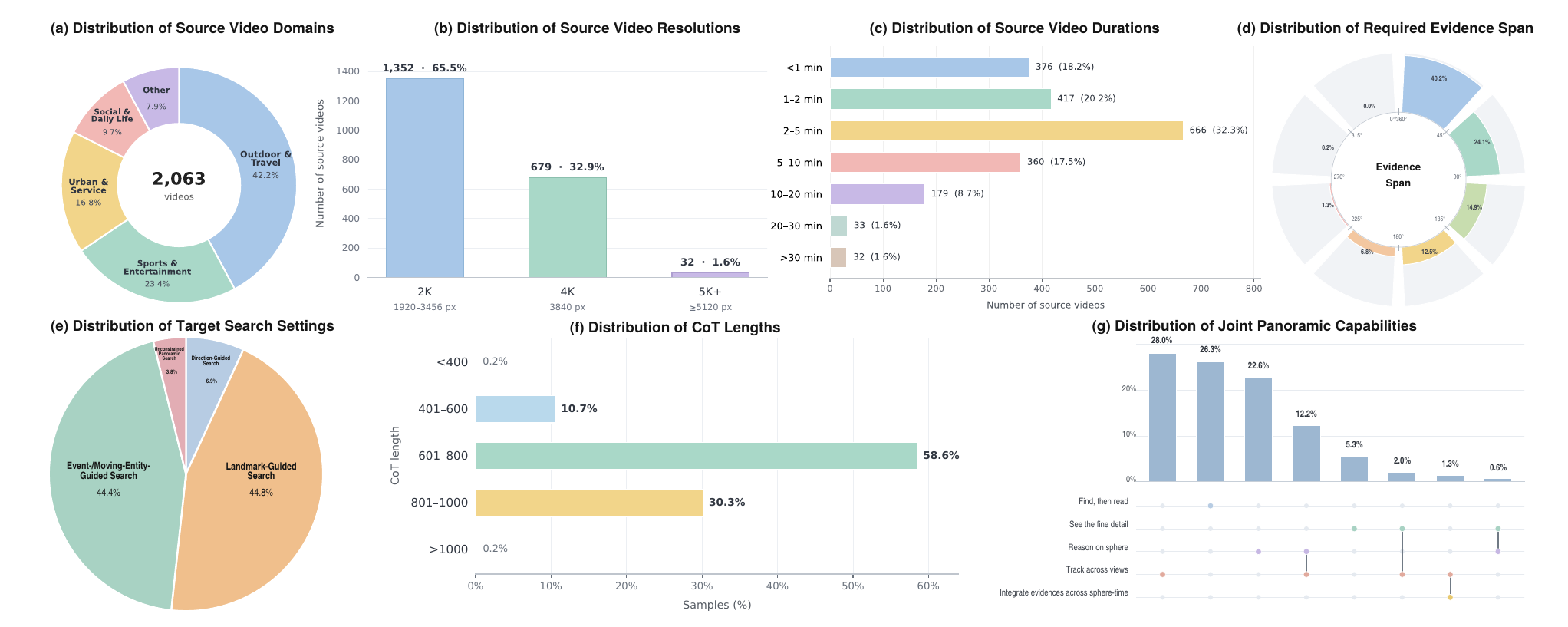}
    \caption{
    \textbf{Statistics of PanoVideoCoT.}
    The source-video plots summarize duration and resolution. The remaining plots report angular evidence span, target-search setting, CoT length, and the co-occurrence of panoramic capabilities. Together they show how the corpus combines visually detailed source videos with supervision that connects evidence across viewing directions and time.
    }
    \label{fig:dataset_statistics}
\end{figure*}

The corpus combines the spatial detail and temporal extent needed for panoramic reasoning. Its sources range from short clips to long-form recordings, and 34.5\% are recorded at 4K resolution or above. High resolution preserves local evidence after perspective reprojection, while longer videos support questions that connect temporally separated events. The corpus averages 6.5 QA--CoT pairs per video, turning each source into several relation-specific problems built from different evidence sets. Approximately 100K evidence nodes make this reuse possible: different questions can select distinct entities, events, directions, and temporal relations from the same video.

The questions emphasize finding and connecting panoramic evidence. Only 6.9\% name the target direction, while 89.2\% rely on a landmark, event, or moving entity and the rest require unconstrained search. The supervision therefore teaches the model to use scene content and ongoing activity to decide where to inspect. Moreover, 59.8\% span more than $45^\circ$, placing most beyond a narrow local view. Track Across Views and Reason on the Sphere appear in 44.4\% and 36.4\% of the corpus, and 68.6\% require at least one. Their 12.2\% intersection combines identity continuity with spherical reasoning.

\subsection{Training Strategy}

\paragraph{Panoramic CoT SFT.} In the first stage, we fine-tune the general video MLLM on PanoVideoCoT. Given the original ERP video and the multiple-choice question, including its options, the model learns to generate the synthesized CoT followed by the answer. The CoT exposes which observations must be connected and in what order, providing direct supervision for cross-view identity, spherical relations, and temporal composition. Generating the answer after the trace makes evidence composition part of the model's decoding path rather than a separate auxiliary target.

\paragraph{Evidence-Privileged On-Policy Self-Distillation (EP-OPSD).} EP-OPSD applies Vision-OPD's On-Policy Self-Distillation to panoramic reasoning while retaining its optimization objective and training configuration~\cite{yuan2026visionopd}. The task-specific adaptation lies in the teacher's privileged visual context. The student samples a trajectory from its current policy using the ERP video, while the teacher follows the same response prefix with additional perspective views from the sample's verified evidence nodes. These views supply the local content and spherical direction needed to guide the student's current partial solution. Panoramic CoT SFT teaches complete solution patterns, and EP-OPSD aligns the student's own reasoning path with their supporting visual evidence.

\section{Experiments}\label{sec:experiments}
\subsection{Experimental Settings}

\paragraph{Benchmark evaluation.} We evaluate eleven baselines: two proprietary models, eight open-source general-purpose models, and one panoramic model. Each receives the original ERP video, sampled at 1 fps for at most 128 frames and resized to a 1280-pixel long edge, without perspective views or directional overlays. We report overall accuracy across five semantic categories, four single-label target-search settings, and five multi-label panoramic capabilities. Capability accuracy uses all questions carrying the corresponding label.

\paragraph{PanoVideoReasoner training.} We initialize PanoVideoReasoner from Qwen3.5-9B and train it with LoRA using rank 8 and $\alpha=32$. Each sample uses the original ERP video, uniformly sampled over its full duration at a target rate of 1 fps and capped at 128 frames, with a decoding width of 1280 pixels. PanoVideoCoT and PanoVideoBench use disjoint source videos. Appendix~\ref{app:reasoner_training} provides the remaining optimization and input-budget settings.

\subsection{Main Results}

Table~\ref{tab:main_results} reveals a consistent gap: current models recognize visible content more reliably than they preserve entities and relations across viewing directions and time. Baseline accuracy falls from 55.0\% on APP to 33.9\% on MOT and 31.9\% on TEM, while the panoramic PanoWorld-Hstar remains comparable to general video MLLMs. Overall scores also conceal distinct failure modes, with similarly ranked models diverging sharply on temporal reasoning. Fig.~\ref{fig:diagnostic_radar} localizes the bottleneck: fine-detail recognition is comparatively strong, but cross-view tracking and sphere--time integration remain weak. Panoramic video understanding therefore depends on building a scene representation that persists as evidence moves out of view and reappears later.

\par\smallskip
\noindent\begin{minipage}{\textwidth}
    \vspace{0pt}
    \centering
    \captionof{table}{
    \textbf{PanoVideoBench comparison.}
    Overall reports accuracy on all 1,000 questions; APP, SPA, ACT, MOT, and TEM report accuracy for the five semantic categories. Models are grouped into proprietary, open-source general-purpose, and open-source panoramic systems. Best results are in \textbf{bold}; second-best are \underline{underlined}.
    }
    \label{tab:main_results}
    \footnotesize
    \setlength{\tabcolsep}{3.8pt}
    \renewcommand{\arraystretch}{0.96}
    \begin{tabular}{@{}lcccccc@{}}
        \toprule
        Model & Overall & APP & SPA & ACT & MOT & TEM \\
        \midrule

        \multicolumn{7}{l}{\textit{Proprietary models}} \\
        Gemini 3.7 Flash~\cite{deepmind2026gemini37flash}
        & \underline{53.1}
        & \textbf{65.7}
        & \underline{49.5}
        & \underline{52.4}
        & \underline{40.9}
        & 36.5 \\

        GPT-5.6-sol~\cite{openai2026gpt56}
        & 49.6
        & \underline{59.9}
        & 46.0
        & 51.9
        & 33.5
        & \underline{41.9} \\

        \midrule
        \multicolumn{7}{l}{\textit{Open-source general-purpose models}} \\
        Gemma-4-31B-it~\cite{gemmateam2026gemma4}
        & 47.0 & 58.4 & 48.5 & 44.6 & 34.8 & 27.0 \\

        Nemotron Omni 30B-A3B~\cite{nvidia2026nemotron3nanoomni}
        & 46.3
        & \underline{59.9}
        & 46.0
        & 41.6
        & 31.1
        & 35.1 \\

        InternVL3.5-8B~\cite{wang2025internvl35}
        & 44.9 & 53.8 & 36.1 & 49.4 & 34.1 & 39.2 \\

        Qwen3.8-27B~\cite{qwen2026qwen38}
        & 44.8 & 54.7 & 44.1 & 46.8 & 31.1 & 27.0 \\

        Qwen3.5-9B~\cite{qwen2026qwen35}
        & 43.2 & 55.4 & 36.6 & 42.5 & 35.4 & 27.0 \\

        Eagle2.5-8B~\cite{chen2025eagle25}
        & 39.6 & 50.5 & 35.6 & 38.6 & 32.3 & 21.6 \\

        MiniCPM-V-4.6~\cite{yao2024minicpm,openbmb2026minicpmv46}
        & 39.0 & 48.9 & 37.1 & 34.3 & 34.1 & 25.7 \\

        GLM-4.1V-9B-Thinking~\cite{glmvteam2025glm41v}
        & 38.0 & 46.5 & 38.6 & 32.2 & 31.1 & 32.4 \\

        \midrule
        \multicolumn{7}{l}{\textit{Open-source panoramic models}} \\
        PanoWorld-Hstar~\cite{wang2026panoworld}
        & 42.9 & 51.7 & 45.0 & 36.5 & 34.1 & 37.8 \\

        \midrule
        \textbf{PanoVideoReasoner}
        & \textbf{55.7}
        & 59.6
        & \textbf{52.5}
        & \textbf{60.1}
        & \textbf{48.8}
        & \textbf{48.6} \\

        \bottomrule
    \end{tabular}
\end{minipage}
\par\smallskip

\noindent\begin{minipage}{\textwidth}
    \vspace{0pt}
    \centering
    \includegraphics[
        width=0.82\linewidth,
        trim=4 4 4 4,
        clip
    ]{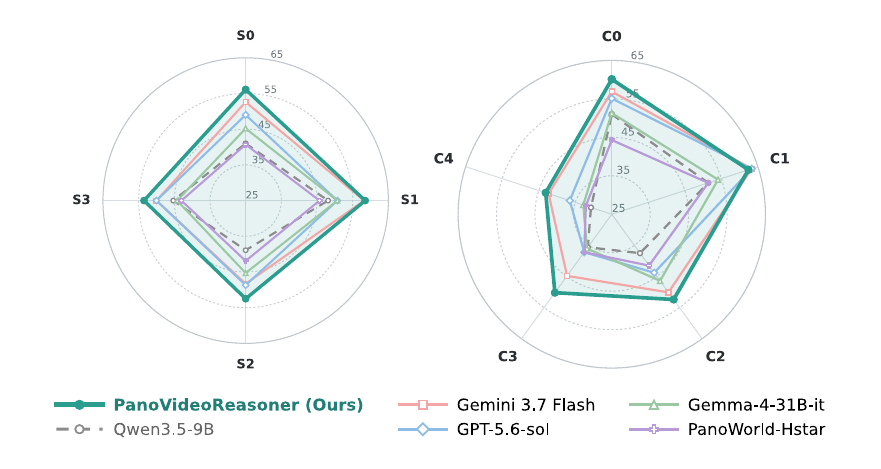}

    \captionof{figure}{
    \textbf{Accuracy across target-search settings and panoramic capabilities.}
    The left radar shows Direction-Guided (S0), Landmark-Guided (S1), Event-/Moving-Entity-Guided (S2), and Unconstrained Panoramic Search (S3). The right radar shows Find, then Read (C0), See the Fine Detail (C1), Reason on the Sphere (C2), Track Across Views (C3), and Integrate Evidence Across Sphere--Time (C4). C0 denotes local readout, whereas C1--C4 can co-occur when a question requires several panoramic operations; each right-hand score uses all questions carrying that label.
    }
    \label{fig:diagnostic_radar}
\end{minipage}
\par\vspace{2pt}

PanoVideoReasoner directly improves this capability profile. It reaches a new SOTA of 55.7\%, 12.5 points above its Qwen3.5-9B backbone, and ranks first on SPA, ACT, MOT, and TEM. APP rises by 4.2 points, while the other four categories improve by 13.4--21.6; the diagnostic gains likewise concentrate on spherical reasoning, cross-view tracking, and sphere--time integration. This pattern shows that the training pipeline strengthens how a general video model connects observations into a persistent scene representation.

\vspace{-7pt}
\paragraph{Effect of input resolution.} Higher resolution makes local evidence easier to read, but its benefit depends on the visual encoder. Table~\ref{tab:input_analysis} shows a clear improvement for Qwen, while Gemma nearly saturates at a 1280-pixel long edge. Gains are most stable on APP and ACT; MOT changes little, and TEM follows opposite trends for the two models. Extra pixels improve per-frame perception, whereas spherical position, identity continuity, and temporal structure must still be learned across observations.

\par\smallskip
\noindent
\begin{minipage}{\textwidth}

\begin{minipage}[t]{0.49\textwidth}
\vspace{0pt}
\captionof{table}{
\textbf{Input analysis on PanoVideoBench.}
(a) varies ERP resolution. (b) compares the default ERP input with an eight-view perspective mosaic (8V) and an orientation-overlaid ERP (Ovl); Appendix~\ref{app:representation_settings} gives their construction and sampling details. Qwen and Gemma abbreviate Qwen3.8-27B and Gemma-4-31B-it; \textbf{Bold} and \underline{underlining} mark each model's best and second best.
}
\label{tab:input_analysis}
\label{tab:input_resolution}
\label{tab:input_representation}
\end{minipage}
\hfill
\begin{minipage}[t]{0.49\textwidth}
\vspace{0pt}
\captionof{table}{
\textbf{Training ablations.}
All variants use the same Qwen3.5-9B backbone and PanoVideoCoT samples. (a) examines first-stage supervision: base, answer-only SFT, and CoT SFT. (b) starts from the CoT SFT checkpoint and compares second-stage OPSD with uniform (Unif.), question-guided (Q-g.), and evidence-privileged (EP) teacher views. \textbf{Bold} and \underline{underlining} mark the top two results.
}
\label{tab:ablations}
\label{tab:ablation_sft}
\label{tab:ablation_epod}
\end{minipage}

\vspace{3pt}

\begin{minipage}[t]{0.24\textwidth}
\vspace{0pt}
\centering
\textbf{(a) ERP resolution}\par\smallskip
\fontsize{5.7}{6.5}\selectfont
\setlength{\tabcolsep}{0.55pt}
\renewcommand{\arraystretch}{1.0}
\begin{tabular}{@{}l*{3}{c}@{\hspace{1.2pt}}*{3}{c}@{}}
\toprule
& \multicolumn{3}{c}{Qwen}
& \multicolumn{3}{c}{Gemma} \\
\cmidrule(lr){2-4}\cmidrule(l){5-7}
Metric & 640 & 1280 & 1920 & 640 & 1280 & 1920 \\
\midrule
Overall
& 42.3 & \underline{44.8} & \textbf{46.2}
& 46.0 & \underline{47.0} & \textbf{47.2} \\
APP
& 46.8 & \underline{54.7} & \textbf{56.0}
& 57.2 & \textbf{58.4} & \underline{58.1} \\
SPA
& \underline{44.6} & 44.1 & \textbf{48.5}
& 47.0 & \textbf{48.5} & \underline{48.0} \\
ACT
& 45.9 & \underline{46.8} & \textbf{48.5}
& 42.9 & \underline{44.6} & \textbf{45.1} \\
MOT
& \underline{31.1} & \underline{31.1} & \textbf{32.9}
& \underline{34.1} & \textbf{34.8} & 32.9 \\
TEM
& \textbf{29.7} & \underline{27.0} & 18.9
& \underline{29.7} & 27.0 & \textbf{35.1} \\
\bottomrule
\end{tabular}
\end{minipage}%
\hfill%
\begin{minipage}[t]{0.24\textwidth}
\vspace{0pt}
\centering
\textbf{(b) Panoramic input}\par\smallskip
\fontsize{5.7}{6.5}\selectfont
\setlength{\tabcolsep}{0.55pt}
\renewcommand{\arraystretch}{1.0}
\begin{tabular}{@{}l*{3}{c}@{\hspace{1.2pt}}*{3}{c}@{}}
\toprule
& \multicolumn{3}{c}{Qwen}
& \multicolumn{3}{c}{Gemma} \\
\cmidrule(lr){2-4}\cmidrule(l){5-7}
Metric & ERP & 8V & Ovl & ERP & 8V & Ovl \\
\midrule
Overall
& \textbf{44.8} & 41.0 & \underline{44.5}
& \textbf{47.0} & 40.9 & \underline{44.5} \\
APP
& \textbf{54.7} & 47.7 & \underline{48.6}
& \textbf{58.4} & 43.7 & \underline{45.6} \\
SPA
& 44.1 & \underline{44.6} & \textbf{48.0}
& \underline{48.5} & 40.6 & \textbf{50.5} \\
ACT
& \textbf{46.8} & 42.5 & \underline{45.1}
& \textbf{44.6} & 42.1 & \underline{43.3} \\
MOT
& \underline{31.1} & 26.8 & \textbf{34.1}
& 34.8 & \underline{37.8} & \textbf{42.1} \\
TEM
& 27.0 & \underline{28.4} & \textbf{37.8}
& \underline{27.0} & \textbf{32.4} & \textbf{32.4} \\
\bottomrule
\end{tabular}
\end{minipage}%
\hfill%
\begin{minipage}[t]{0.235\textwidth}
\vspace{0pt}
\centering
\textbf{(a) SFT target}\par\smallskip
\fontsize{6.0}{7.0}\selectfont
\setlength{\tabcolsep}{3.0pt}
\renewcommand{\arraystretch}{1.16}
\begin{tabular}{@{}lccc@{}}
\toprule
Metric & Base & Answer & CoT \\
\midrule
Overall & 43.2 & \underline{45.6} & \textbf{51.8} \\
APP     & \underline{55.4} & 55.0 & \textbf{57.2} \\
SPA     & 36.6 & \underline{41.1} & \textbf{45.0} \\
ACT     & 42.5 & \underline{45.5} & \textbf{57.1} \\
MOT     & 35.4 & \underline{36.0} & \textbf{45.1} \\
TEM     & 27.0 & \underline{37.8} & \textbf{44.6} \\
\bottomrule
\end{tabular}
\end{minipage}%
\hfill%
\begin{minipage}[t]{0.275\textwidth}
\vspace{0pt}
\centering
\textbf{(b) Teacher views}\par\smallskip
\fontsize{5.9}{7.0}\selectfont
\setlength{\tabcolsep}{3.0pt}
\renewcommand{\arraystretch}{1.16}
\begin{tabular}{@{}lcccc@{}}
\toprule
Metric & CoT & Unif. & Q-g. & EP \\
\midrule
Overall & 51.8 & 50.9 & \underline{53.9} & \textbf{55.7} \\
APP     & 57.2 & 57.8 & \underline{58.4} & \textbf{59.6} \\
SPA     & 45.0 & 46.5 & \underline{49.0} & \textbf{52.5} \\
ACT     & 57.1 & 55.4 & \underline{58.8} & \textbf{60.1} \\
MOT     & 45.1 & 41.5 & \underline{47.0} & \textbf{48.8} \\
TEM     & 44.6 & 39.2 & \underline{47.3} & \textbf{48.6} \\
\bottomrule
\end{tabular}
\end{minipage}

\end{minipage}
\par\smallskip
\enlargethispage{3\baselineskip}

\paragraph{Effect of panoramic representation.} The 8-view mosaic rectifies each ERP frame into eight perspective views on a compass grid; the orientation overlay preserves the ERP and adds fixed yaw and pitch markers. The mosaic offers familiar local geometry but performs worse overall because the model must reconstruct correspondence and spherical adjacency across tiles. The overlay gives observations a stable spherical address and consistently improves SPA, MOT, and TEM, showing that direction also supports trajectory and event association across views and time. Its weaker APP and ACT results indicate that coordinates cannot replace clean local evidence. Effective panoramic input must preserve global spherical organization while exposing local detail when needed.

\subsection{Ablation Studies}\label{sec:ablation}

\paragraph{CoT supervision in SFT.}
On the same QA pairs, answer-only SFT yields a modest improvement, whereas CoT SFT adds a further 6.2 points (Table~\ref{tab:ablation_sft}a). The gain concentrates on ACT, MOT, and TEM and remains much smaller on APP. Answer labels specify the endpoint; synthesized trajectories additionally supervise how actors, identities, and events persist across changing views and times. CoT SFT consequently improves the same relational capabilities that the benchmark identifies as the main weakness of general video models.

\paragraph{Evidence selection in OPSD.}
Starting from the same CoT SFT checkpoint, we vary the teacher's additional views during OPSD (Table~\ref{tab:ablation_epod}b). Uniform views reduce overall performance and hurt MOT and TEM most, suggesting that extra crops can dilute cross-frame correspondence. Question-guided selection recovers the loss, while verified evidence views perform best in every category and add 3.9 points over CoT SFT. This progression separates coverage, relevance, and evidence validity: verified nodes bind each supporting observation to its time, direction, and cross-view relations. CoT SFT teaches the evidence-chain structure, and EP-OPSD grounds the student's intermediate states along that chain.

\section{Conclusion}

We presented PanoVideoBench, the first manually annotated benchmark for comprehensive panoramic video understanding. Its evaluation identifies a central weakness of current MLLMs: preserving entities and their relations as evidence shifts across viewing directions and time. Guided by this diagnosis, we constructed PanoVideoCoT from real panoramic videos and trained PanoVideoReasoner with Panoramic CoT SFT and EP-OPSD. PanoVideoReasoner reaches 55.7\% accuracy, outperforming its backbone by 12.5 points and the previous SOTA by 2.6 points, with the largest gains in spatial, cross-view, and temporal reasoning. Together, PanoVideoBench and PanoVideoCoT provide the evaluation and supervision needed to develop general models that reason over the full sphere and throughout a video.

\clearpage

\subsection*{Reproducibility Statement}
Sections~\ref{sec:benchmark}--\ref{sec:experiments} describe benchmark construction, sphere--time-grounded synthesis, training, and evaluation. The appendix provides the full taxonomy and quality-control procedure (Appendix~\ref{app:benchmark_details}), synthesis and trajectory-construction details (Appendix~\ref{app:cot_details}), and prompts, hyperparameters, input processing, and evaluation protocols (Appendix~\ref{app:training_details}). Preview examples from PanoVideoBench and PanoVideoCoT are included with the supplementary materials. The complete benchmark, training corpus, and code will be released upon acceptance.

\subsection*{AI Use Statement}
Generative AI tools assisted the language-only screening of PanoVideoBench questions and answer choices. They were also used to synthesize and refine candidate questions, answer choices, and reasoning traces for the PanoVideoCoT training corpus from structured sphere--time evidence. Human annotators created and reviewed the benchmark against the panoramic videos and made the final decisions on question inclusion and gold answers. The authors reviewed the resulting benchmark, training data, manuscript claims, and artifacts and take full responsibility for the work.

\bibliography{iclr2027_conference}
\bibliographystyle{iclr2027_conference}

\clearpage
\appendix
\section{PanoVideoBench Details}
\label{app:benchmark_details}

\subsection{Quality Control}\label{app:quality_control}

The annotation pool contains 1,296 questions that pass the language-only screening described below. Each item is independently reviewed by two annotators using the full panoramic video. Together with the initial answer label, their judgments form three votes. Of these questions, 983 receive unanimous votes and enter the final benchmark directly. Another 171 receive a two-to-one decision, and their answer labels are corrected to the majority choice. The remaining 142 fail to establish consensus and are discarded. A final adjudication ranks the 171 majority-resolved questions by visual grounding, quality, and difficulty, retaining 17 to complete the 1,000-question benchmark.

Questions involving spatial direction, subject orientation, or viewpoint changes receive an additional reference-frame check. Reviewers use the directional indicators in a spherical video player to re-examine left--right, front--back, up--down, and rotational relations, while explicitly distinguishing observer-centric directions from the intrinsic orientation of scene entities. For cross-view motion and identity-tracking questions, reviewers follow the target continuously along the timeline to verify that observations across different times and viewing directions correspond to the same entity. Counting questions are checked according to the operation required by the question, including the number of static instances at a single time, identity-deduplicated entities accumulated over time, repeated event occurrences, and sets of attributes or categories collected across timestamps. These checks ensure that both the semantic category and the gold answer are grounded in the same visual evidence.

To reduce non-visual shortcuts, GPT-5.4-mini receives only the question and answer options, without access to the video, and flags items that can be solved reliably from language alone. Deterministic checks further remove malformed options, invalid answer labels, and duplicate or near-duplicate questions. We also verify that option shuffling preserves the answer mapping and balance the final distribution of correct-option positions across A--D.
\subsection{Question Taxonomy}
\label{app:question_taxonomy}

Each question is assigned according to the visual operation that determines its answer. The five primary categories define the scope of panoramic video understanding, while the 15 secondary types distinguish the operations within that scope. The definitions below also clarify boundaries between types that may draw on similar evidence.

\paragraph{Panoramic Appearance and Detail Perception (APP).} APP contains questions whose answers are determined by the visible appearance of the panoramic scene. The central challenge is to preserve the complete surrounding context while retaining enough local detail to distinguish the answer choices. Equirectangular projection introduces position-dependent distortion, whereas a local perspective view improves geometry and effective resolution at the cost of excluding other directions.

\textbf{Scene and Layout Understanding (APP-SCN).} These questions integrate scene information distributed across multiple viewing directions into a single representation of the environment. The answer may describe the scene type, the division of major regions, or the overall layout among large structures. When a sufficiently wide local view determines a named place or facility, the question belongs to APP-ENT. Questions that locate the observer or camera within the environment belong to SPA-OBS.

\textbf{Entity and Attribute Recognition (APP-ENT).} The decisive operation is to find the relevant entity in the full panorama and obtain enough entity-level evidence to distinguish the choices. This type covers entity category, color, clothing, presence, and other visible attributes. It also includes static instance counting, where the answer can be obtained from the entities visible at the relevant moment without associating identities across time.

\textbf{Fine-Grained Visual Recognition (APP-FGR).} These questions depend on small, subtle, or easily confused local cues, including text, digits, symbols, signs, logos, and fine patterns. A question is assigned to APP-FGR when the choices cannot be distinguished without resolving the fine detail. If the detail appears only in the question as a cue for locating another target, it does not determine the question type.

\paragraph{Spherical Spatial and Viewpoint Reasoning (SPA).} SPA contains questions determined by position, direction, orientation, or reference-frame relations in the complete viewing sphere. The answer requires observations to be placed in a shared spatial frame rather than interpreted as unrelated local crops.

\textbf{Observer-Centric Spatial Relations (SPA-OBS).} The observer or camera is the key reference object or the object being located. This type covers azimuthal relations such as left, right, front, rear, diagonal directions, clockwise order, and the number of entities lying between two targets around the observer. It also covers elevation and observer-location questions, including whether a target lies above or below the camera, the camera's height relative to a person, the floor on which the camera is located, whether it is indoors or outdoors, and the platform on which it is mounted. The decisive operation is to map targets from different local views into one observer-centered spatial frame, or to determine the observer's own position in the scene. A relation between targets in different viewing directions remains SPA-OBS when the answer is their observer-centered order. It becomes SPA-ORI when an actual or hypothetical change of orientation must be applied before the relation can be determined.

\textbf{Object-Centric Spatial Relations (SPA-OBJ).} A scene entity, object, or landmark serves as the spatial anchor. The answer describes where target A lies relative to anchor B, rather than where the target lies relative to the observer. When the anchor only indicates where to look and the choices differ in what object or attribute is present there, recognition determines the answer and the question belongs to APP.

\textbf{Orientation and Viewpoint Reasoning (SPA-ORI).} These questions ask which direction an observer or scene entity faces, or how a direction changes under an actual or hypothetical viewpoint transformation. The type includes static orientation, the distinction between the observer's side and another entity's own side, direction after a person or camera turns, and counterfactual rotations such as a $90^\circ$ change in viewpoint. SPA-OBS and SPA-OBJ answer where a target is, whereas SPA-ORI answers which way it faces or how its direction transforms. A turn belongs to SPA-ORI when the transformed direction determines the answer. Questions that instead attribute an observed change to camera motion, object motion, or both belong to MOT-COM.

\paragraph{Omnidirectional Action and State Understanding (ACT).} ACT contains questions whose answers are determined by an action, interaction, state, or state change. Unlike video recorded with a preselected field of view, panoramic video does not keep the relevant actor or interaction centered, so understanding may require following activity across viewing directions.

\textbf{Action and Interaction Recognition (ACT-INT).} The answer identifies what action or interaction occurs. The decisive evidence is the activity itself, rather than the state produced after the activity finishes.

\textbf{State and Transition Understanding (ACT-STA).} The answer describes an entity's current state, whether a state persists, or how it changes from one state to another. Questions are assigned here when the answer variable is the state. If the choices instead express when a state occurs, which state comes first, whether two states overlap, or how long one lasts, the answer is temporal and the question belongs to TEM.

\textbf{Multi-Entity Activity Understanding (ACT-MUL).} The answer requires the activities of multiple entities to be understood jointly, including concurrent, coordinated, or otherwise related behavior. The mere presence of several entities does not make a question ACT-MUL. Their activities must jointly determine the answer.

\paragraph{Cross-View Motion and Identity Tracking (MOT).} MOT contains questions determined by an entity's motion over time, its trajectory across viewing directions, its identity continuity, or the relation between camera motion and object motion. The relevant evidence forms a continuous motion or identity chain rather than a set of independent observations.

\textbf{Trajectory and Identity Tracking (MOT-TRK).} The decisive operation is to follow a target over time and across viewing directions while preserving its identity. This operation supports recovery of motion direction, path, starting or ending position, and cross-view trajectory. It also covers identity-aware counting: repeated observations must first be associated with the same entity and then deduplicated to obtain the number of distinct entities. Tracking and identity continuity are treated as two parts of the same process.

\textbf{Camera--Object Motion Disambiguation (MOT-COM).} These questions separate an observed change in position, direction, or relative geometry into camera motion, object motion, or their joint effect. Apparent displacement in the video cannot be interpreted directly as physical motion of the object. The answer must identify which component produced the observed change.

\paragraph{Panoramic Spatio-Temporal Event Reasoning (TEM).} TEM contains questions that locate, order, count, or aggregate events across the joint spherical and temporal extent of a video. Multiple temporal positions may be involved in each case, but the subtype follows what the answer asserts: a temporal location or duration, a temporal relation, a number of event occurrences, or a result aggregated across time.

\textbf{Event Localization (TEM-LOC).} The event must first be recognized and then placed on the timeline. The answer is a timestamp, start or end boundary, temporal interval, or duration.

\textbf{Event Ordering (TEM-ORD).} The answer expresses a temporal relation between events, such as which occurs first, whether they occur simultaneously, their ordinal position, or their complete sequence. An event mentioned in the question may simply indicate when to inspect the video. If the choices at that moment differ in action, pose, object, or state, that content determines the question type rather than TEM-ORD.

\textbf{Event Counting (TEM-COUNT).} The decisive operation is to identify every occurrence of a target event and accumulate the total. The same entity may contribute several event instances. Repeated transient events, such as multiple fireworks, also belong to this type because the answer counts occurrences rather than distinct identities.

\textbf{Cross-Time Evidence Aggregation (TEM-AGG).} The answer must combine information collected from multiple temporal locations. This includes forming a set across time and counting its distinct attributes or categories, comparing observations from different intervals, and producing a joint description whose components occur at separate times. Every contributing interval must be necessary to distinguish the correct answer. The defining operation is aggregation across time, not simply a large temporal gap. Static instance counts belong to APP-ENT, identity-deduplicated entity counts to MOT-TRK, event occurrences to TEM-COUNT, and sets accumulated across time to TEM-AGG.

\subsection{Diagnostic Annotation Details}

The two diagnostic axes describe how the supporting evidence is reached and what must be done with it. Each question receives one target-search label. A timestamp or interval may narrow when to look, but the search label is determined by the spatial cue that remains.

\paragraph{Direction-Guided Search (S0).} The question directly names the relevant viewing direction through an observer-centered expression such as front, rear, left, right, above, below, or a rotational offset. The target view is specified, although answering the question may still require tracking or integrating evidence after that view is reached.

\paragraph{Landmark-Guided Search (S1).} A stationary and visually persistent scene element identifies the relevant part of the panorama. The model must first find this landmark and then inspect the target at, near, or relative to it. When the spatial cue is an action or an entity that changes position over time, the search is categorized as S2 instead.

\paragraph{Event-/Moving-Entity-Guided Search (S2).} The relevant view is identified by an event, action, state change, or moving entity. Reaching the evidence requires detecting the occurrence or following the entity as its viewing direction changes. This label describes how the evidence is found; the question type is determined separately by the answer being sought.

\paragraph{Unconstrained Panoramic Search (S3).} The question provides no viewing direction, static landmark, or dynamic spatial anchor. It specifies what information is needed but leaves its position open, requiring a search of the surrounding sphere over the relevant temporal range.

Capability labels describe the visual operations needed to turn the retrieved evidence into an answer. C0 is used when local readout is sufficient, while C1--C4 can co-occur when a question combines several panoramic operations.

\paragraph{Find, then Read (C0).} The decisive evidence can be read directly from a single local view once that view has been found. The panoramic challenge lies in reaching the target region rather than resolving fine detail, transforming directions, maintaining cross-view identity, or combining several observations.

\paragraph{See the Fine Detail (C1).} The answer depends on a small, distorted, or low-effective-resolution cue, such as text, a digit, a symbol, a logo, or a subtle visual attribute. The label applies when resolving that cue is necessary to distinguish the answer choices, not when the detail merely helps locate another target.

\paragraph{Reason on the Sphere (C2).} The evidence must be interpreted in a common spherical or viewpoint-centered reference frame. This includes azimuth and elevation, clockwise order, relations between targets in different viewing directions, facing orientation, and directional changes under actual or hypothetical viewpoint transformations.

\paragraph{Track Across Views (C3).} The answer depends on maintaining correspondence as an entity leaves one local view, enters another, or reappears later. The operation supports identity continuity, trajectory recovery, and identity-aware counting across changes in time and viewing direction. Static relations between different directions are covered by C2, while C3 requires correspondence over a sequence of observations.

\paragraph{Integrate Evidence Across Sphere--Time (C4).} Two or more observations from different directions or temporal locations are jointly necessary for the answer. C4 covers comparison, ordering, aggregation, and joint descriptions for which no single observation is sufficient. It can co-occur with C2 or C3 when the evidence must also be mapped on the sphere or linked by identity.

\section{PanoVideoCoT Construction Details}
\label{app:cot_details}
Each QA--CoT pair is backed by a stored evidence record and an accepted exploration trajectory. Every model-based role uses Qwen3.8-27B~\cite{qwen2026qwen38}.

\subsection{Evidence Record and Question Refinement}

\paragraph{Stored evidence.} An entity node contains a stable visual description and a sequence of observations. Each observation stores an appearance, action, event, or state together with its timestamp or interval, yaw, pitch, and perspective view. Cross-view and cross-clip associations are verified from appearance, scene context, and feasible motion before their observations are joined. Only timestamped observations enter the final record. Each question stores the observations supporting its answer together with AES and the two diagnostic annotations.

\paragraph{Critic cycle.} Each refinement round runs semantic, visual, blind, and language-quality review in sequence. The semantic critic checks the selected evidence relation and answer uniqueness. The visual critic combines sparse global frames, dense evidence intervals, and targeted perspective views to verify entities, timing, spherical relations, cross-view identity, and all four options. The blind critic evaluates shuffled questions and options without video, while the language-quality critic improves clarity and option parallelism. A critic returns the affected field and its evidence, allowing the refiner to update the question, options, or evidence interval before the next round. The final joint audit evaluates the complete QA pair against the video.

\subsection{Trajectory Construction and Certification}

\paragraph{Two-stage exploration.} Blind exploration starts from uniformly sampled ERP overviews. The agent can resample a complete panorama at a chosen time or request a perspective view at a selected time and direction. The exploration prompt encourages the agent to inspect the supporting sphere-time region before answering. Each stage stops at the first rollout that selects the correct option, with at most four rollouts. Questions not solved in the blind stage enter evidence-guided exploration, which adds textual descriptions of the verified observations with their time, yaw, and pitch. These cues guide further visual inspection without exposing evidence views or the answer. Questions still unsolved after four guided rollouts do not enter the corpus.

\paragraph{CoT reconstruction.} We merge repeated inspections at the same time and direction, then distill the trajectory into its search goal, decisive observations, revised expectations, and option eliminations. The writer composes these elements into a continuous CoT, and an attribution model links each visual statement to its executed inspection. Certification checks whether the CoT body supports its annotated answer and whether every attributed statement agrees with the time and direction of the corresponding inspection. The first CoT that passes these checks and the joint option audit is retained.

\section{Training and Evaluation Details}
\label{app:training_details}

\subsection{PanoVideoReasoner Training}\label{app:reasoner_training}

\paragraph{Panoramic CoT SFT.} We fine-tune Qwen3.5-9B for one epoch with LoRA using rank 8, $\alpha=32$, and zero dropout. The peak learning rate is $1\times10^{-4}$ under a cosine schedule with 200 warmup steps. We use a weight decay of 0.01 and clip the gradient norm at 1.0.

\paragraph{Video input.} Training uses the original ERP video. Frames are sampled uniformly over the full duration at a target rate of 1 fps, capped at 128 frames, and decoded at a width of 1280 pixels. The pixel budget is 52,428,800 per video and 409,600 per frame.

\paragraph{Evidence-privileged OPSD.} Both student and teacher are initialized from the Qwen3.5-9B checkpoint after Panoramic CoT SFT. The student receives the ERP video and samples eight on-policy rollouts per question. We use the recommended Qwen3.5 thinking-mode configuration~\cite{qwen2026qwen35}: temperature 1.0, top-$p$ 0.95, top-$k$ 20, minimum-$p$ 0, presence penalty 1.5, and repetition penalty 1.0. Each rollout is capped at 8,192 tokens. For the same student-generated prefix, the teacher additionally receives up to 12 perspective views selected from the sample's verified evidence nodes. We use all available views when there are at most 12; otherwise, 12 are selected evenly along the time-ordered evidence sequence. Each view is rendered at $448\times448$ with a $67.5^\circ$ field of view. Following Vision-OPD~\cite{yuan2026visionopd}, we minimize generalized Jensen--Shannon divergence with $\beta=0.5$ over the teacher's top-100 tokens together with a tail-probability bucket, and clip the token-level importance ratio at 2.0. Gradients update the student, while the teacher is updated by exponential moving average with a 0.05 contribution from the current student. Training is sequential: SFT provides the initialization, and EP-OPSD optimizes the distillation objective for one epoch with a batch size of 96, a learning rate of $2\times10^{-6}$, and 10 warmup steps.

\subsection{Benchmark Evaluation Protocol}

\paragraph{Video input and decoding.} Each model receives the original ERP video without additional perspective views or orientation overlays. Frames are sampled uniformly over the full duration at a target rate of 1 fps, capped at 128 frames, and resized to a 1280-pixel long edge. We retain the temperature specified by each model's default generation configuration and allow up to 32,768 output tokens, including thinking tokens.

All models are evaluated with the same system and user prompts below.

\begin{lstlisting}[style=synthesisprompt]
SYSTEM
You are watching a 360-degree panoramic video. The video is represented in
equirectangular projection (ERP), where the left and right image boundaries
are spatially continuous.

Inspect the complete video across time and all viewing directions. Track
relevant objects and people consistently across frames. Interpret spatial
directions relative to the observer or the reference object specified in the
question.

USER
The complete input is a 360-degree panoramic video lasting approximately
{duration:.1f} seconds.

{question}
A. {option_a}
B. {option_b}
C. {option_c}
D. {option_d}
\end{lstlisting}

\subsection{Panoramic Representation Settings}
\label{app:representation_settings}

All representation comparisons use the same temporal sampling and input size as the default ERP setting: uniform sampling over the full video at a target rate of 1 fps, at most 128 frames, and a 1280-pixel final long edge.

\paragraph{ERP resolution.} The resolution study keeps the original $2{:}1$ ERP representation and varies its long edge among 640, 1280, and 1920 pixels. Frames are sampled uniformly over the complete video at a target rate of 1 fps, with at most 128 frames.

\paragraph{Compass-8 perspective mosaic.} For every uniformly sampled ERP frame, we render eight tangent-plane perspective views on the equator. Their yaw angles are $-135^\circ$, $-90^\circ$, $-45^\circ$, $0^\circ$, $45^\circ$, $90^\circ$, $135^\circ$, and $180^\circ$; pitch is fixed at $0^\circ$, and both horizontal and vertical fields of view are $90^\circ$. Consecutive view centers are separated by $45^\circ$, producing $45^\circ$ overlap. The frame-level mosaic follows the compass arrangement below.
\begin{center}
\begin{tikzpicture}[
  view/.style={draw=ExampleRule, fill=ExampleQuestionWash, rounded corners=2pt,
    minimum width=0.285\textwidth, minimum height=0.74cm, align=center,
    inner sep=4pt, font=\sffamily\small, text=ExampleInk},
  empty/.style={draw=ExampleRule, densely dashed, fill=white, rounded corners=2pt,
    minimum width=0.285\textwidth, minimum height=0.74cm}
]
\node[view]  at (-4.85, 0.94) {\textbf{Front-left}\\[-1pt]{\scriptsize yaw $-45^\circ$}};
\node[view]  at ( 0.00, 0.94) {\textbf{Front}\\[-1pt]{\scriptsize yaw $0^\circ$}};
\node[view]  at ( 4.85, 0.94) {\textbf{Front-right}\\[-1pt]{\scriptsize yaw $45^\circ$}};
\node[view]  at (-4.85, 0.00) {\textbf{Left}\\[-1pt]{\scriptsize yaw $-90^\circ$}};
\node[empty] at ( 0.00, 0.00) {};
\node[view]  at ( 4.85, 0.00) {\textbf{Right}\\[-1pt]{\scriptsize yaw $90^\circ$}};
\node[view]  at (-4.85,-0.94) {\textbf{Back-left}\\[-1pt]{\scriptsize yaw $-135^\circ$}};
\node[view]  at ( 0.00,-0.94) {\textbf{Back}\\[-1pt]{\scriptsize yaw $180^\circ$}};
\node[view]  at ( 4.85,-0.94) {\textbf{Back-right}\\[-1pt]{\scriptsize yaw $135^\circ$}};
\end{tikzpicture}
\vspace{-2pt}

{\footnotesize Each tile has pitch $0^\circ$ and a $90^\circ\!\times90^\circ$ field of view. The center tile remains empty.}
\end{center}

\paragraph{ERP with orientation overlay.} This setting retains the uncropped $2{:}1$ ERP and adds a fixed, question-independent spherical orientation grid without perspective projection. Yellow dashed meridians mark yaw at $-180^\circ$, $-135^\circ$, $-90^\circ$, $-45^\circ$, $0^\circ$, $45^\circ$, $90^\circ$, $135^\circ$, and $180^\circ$, labeled B, BL, L, FL, F, FR, R, BR, and B along the top. Cyan dashed latitudes mark pitch at $45^\circ$, $0^\circ$, and $-45^\circ$, with their values shown at the right edge. Labeling both ERP boundaries as B makes their spherical continuity explicit. The overlay contains no timestamps, target boxes, evidence locations, or answer-dependent information.

\section{Qualitative Reasoning Examples}
\label{app:qualitative_examples}

Figures~\ref{fig:qual-motion-orientation}--\ref{fig:qual-cross-window} compare PanoVideoReasoner with the Qwen3.8-27B baseline on three complementary reasoning demands. The first follows body orientation and motion through a turn. The second confirms a small emblem across repeated observations at different times and panoramic positions. The third keeps two subjects and two distant time windows separate before composing their destinations. This final comparison also shows that a longer trace does not by itself resolve the association problem: PanoVideoReasoner answers in 189 tokens, whereas Qwen3.8-27B reaches 16K tokens without a valid answer under a nearly matched visual-token budget.

\clearpage
\begin{QualitativeExample}
\ExamplePanel{Question}{%
  {\small\bfseries Between 50 and 60 seconds, what is the final orientation and movement direction of the man in the grey hoodie after he turns around?}\par
  \vspace{4pt}%
  \ExampleOptions
    {Facing away from the camera, moving towards the staircase on the left}
    {Facing the camera, moving towards the staircase on the right}
    {Facing away from the camera, moving towards the staircase on the right}
    {Facing the camera, moving towards the staircase on the left}}

\vspace{4pt}
\ExampleEvidencePanel{%
  \ExampleTimeline{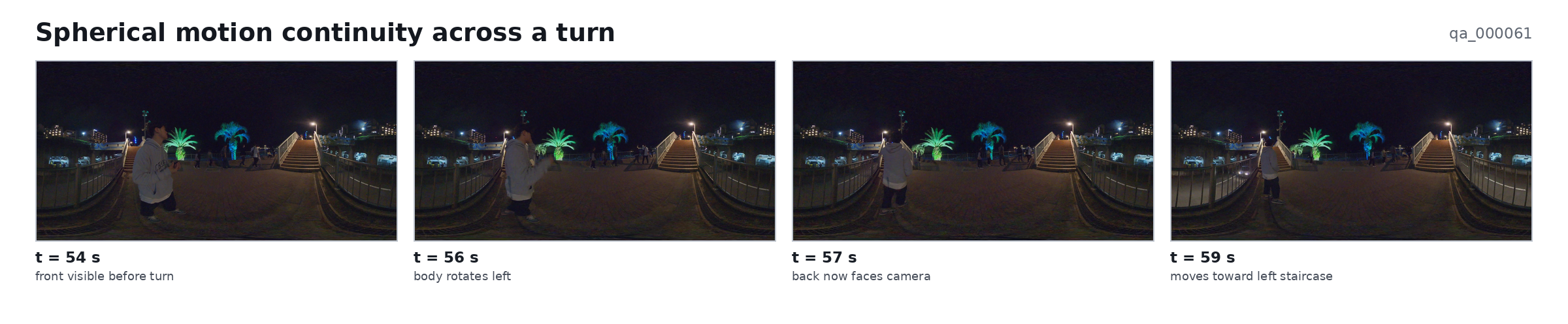}}

\vspace{4pt}
\ExampleReasoning{At around 51 seconds, looking toward the rear of the scene, a man in a grey hoodie is walking across the paved plaza. His back is turned to the camera and his body is angled toward the illuminated staircase set on that rear/left side, with railings and steps clearly ahead of him. From about 53 to 57 seconds he continues turning away from the viewpoint rather than facing back toward us; by roughly 58 to 60 seconds his back is still presented to the camera and he is advancing along the path that leads up to the left-side stairs. There is no clear shift toward the right-hand staircase during this interval, so the motion should not be read as heading to the other stairway. The final orientation is therefore with his back to the camera, moving toward the staircase on the left.}

\ExampleComparison
  {A. Facing away from the camera and moving toward the left staircase.}
  {B. Facing the camera and moving toward the right staircase.}

\ExampleCaption{\textbf{Maintaining motion and orientation through a turn.} The sequence follows the same person from a frontal view through the turn until his back faces the camera and he moves toward the left staircase. PanoVideoReasoner selects A, whereas Qwen3.8-27B reverses both the final orientation and destination.}{fig:qual-motion-orientation}
\end{QualitativeExample}

\clearpage
\begin{QualitativeExample}
\ExamplePanel{Question}{%
  {\small\bfseries When the man in the dark blue varsity jacket and sunglasses turns to face the camera, what letter is visible on the orange logo on his chest?}\par
  \vspace{4pt}%
  \ExampleOptions
    {The letter P}
    {The letter B}
    {The letter D}
    {The letter O}}

\vspace{4pt}
\ExampleEvidencePanel{%
  \ExampleTimeline{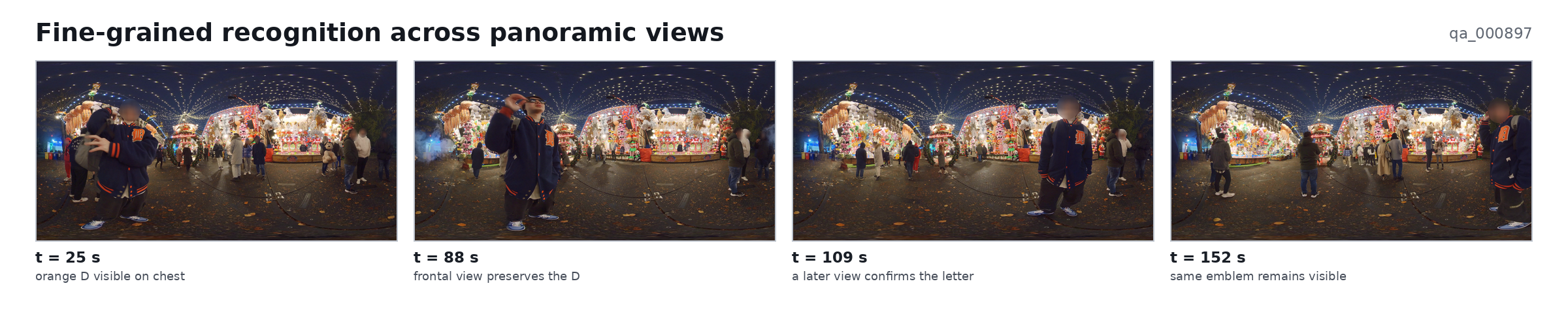}}

\vspace{4pt}
\ExampleReasoning{At the start of the clip, looking toward the rear-left, the man in the dark blue varsity jacket and sunglasses is bent forward with his back to the camera, so only his jacket sleeve and the gray shoulder panel are visible. Around 25 seconds he straightens up, turns slightly, and by about 30 seconds he is more upright and angled toward the center of the view; shortly after he faces the camera again around 80 seconds and stays mostly in front. When his body is turned so the chest is visible --- for instance near 81 to 96 seconds, and clearly on both sides around 145 to 156 seconds when he faces right-to-left --- a large orange logo is on the left side of his navy chest. That logo has two closed curves meeting at a point, the upper-right loop closing downward and the lower-left loop curving inward, which is the shape of a D rather than an open P, an O, or a B. The same orange mark repeats on both sides of the stitched panorama, confirming it is his own chest emblem. So the visible letter is D.}

\ExampleComparison
  {C. The letter D.}
  {B. The letter B.}

\ExampleCaption{\textbf{Confirming a fine detail across panoramic views.} The orange chest emblem appears at several times and different ERP positions as the wearer turns. PanoVideoReasoner identifies the repeated mark as D and selects C; Qwen3.8-27B reads it as B at all evaluated input resolutions.}{fig:qual-cross-view-detail}
\end{QualitativeExample}

\clearpage
\begin{QualitativeExample}
\ExamplePanel{Question}{%
  {\small\bfseries First, observe the group walking toward the building entrance between 2:25 and 2:35. Then, observe the pedestrian in a grey-and-white camouflage jacket between 3:05 and 3:15. What are their final destinations?}\par
  \vspace{4pt}%
  \ExampleOptions
    {Both enter the building}
    {The first group enters; the second remains in front of the entrance}
    {Both remain in front of the entrance}
    {The first group leaves the plaza; the second enters the building}}

\vspace{4pt}
\ExampleEvidencePanel{%
  \ExampleTimeline{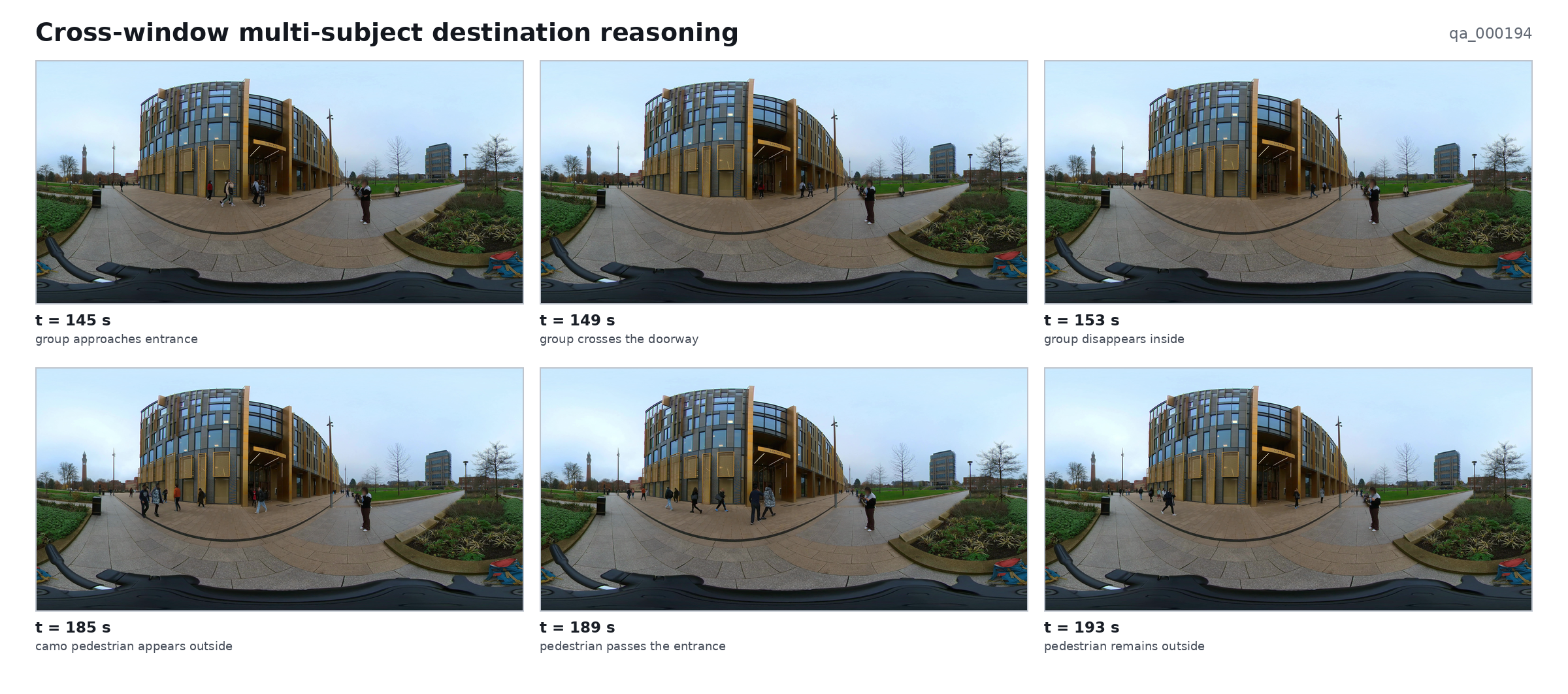}}

\vspace{4pt}
\ExampleReasoning{Looking to the left side of the plaza toward the curved glass-and-wood entrance, the group visible from about 2:25 to 2:35 keeps moving in that direction. By around 2:30 they have reached the doorway and disappear into it rather than turning back or lingering on the open pavement, so their final destination is the building itself. Later, around 3:06, a different pedestrian wearing a grey-and-white camouflage-patterned jacket walks across the foreground and stops just outside the covered entrance; through the following seconds up to about 3:17 they remain standing on the paved area in front of the doors, not crossing inside. The two subjects therefore have different outcomes: the first group enters the building while the camo-jacket pedestrian stays in front of the entrance.}

\ExampleComparison
  {B. The first group enters; the second remains outside.}
  {D. The first group leaves the plaza; the second enters the building.}

\ExampleCaption{\textbf{Composing outcomes across two distant time windows.} The upper sequence tracks the first group through the doorway at 145--153 seconds; the lower sequence shows a different pedestrian remaining outside at 185--193 seconds.}{fig:qual-cross-window}
\end{QualitativeExample}

\clearpage
\section{Prompt Templates}
\label{app:synthesis_prompts}

All synthesis roles use Qwen3.8-27B. The templates preserve each role's inputs and decision criteria while consolidating repeated formatting instructions. Braced fields are populated with sample-specific evidence, views, questions, options, or trajectories.

\subsection{Building the Sphere-Time Evidence Memory}

\subsubsection{Directional observation}

\begin{lstlisting}[style=synthesisprompt]
SYSTEM
You inspect one fixed viewing direction of a panoramic video. Describe only
content supported by the supplied frames.

USER
Viewing direction: {sector}
Frame times: {times}

Select entities that can anchor a question rather than listing the whole scene.
An entity should be recognizable again from its description, persist across
several frames or undergo an observable change, and remain distinct from the
background. Include a briefly visible entity only when its identity and state
are still clear enough to support a question.

Describe intrinsic cues such as color, clothing, markings, shape, or carried
objects. Do not define an entity by its current image position or viewing
direction, since these change across the sphere. Exclude the vehicle, mount,
pole, or other parts of the recording rig.

For each entity, return:
- local_id: identifier within this directional clip;
- description: visual cues for recognizing it again;
- seen_at: timestamps from the supplied list at which it is clearly visible;
- observation: what it looks like, does, or is locally related to;
- motion: moving only if its position changes in the scene; articulation by a
  stationary entity remains stationary;
- confidence: high when identity is clear across frames, otherwise low.

OUTPUT
{"entities": [{"local_id": "...", "description": "...",
"seen_at": [...], "observation": "...", "motion": "...",
"confidence": "..."}]}
\end{lstlisting}

\subsubsection{Entity association across sphere-time}

\begin{lstlisting}[style=synthesisprompt]
SYSTEM
Determine whether two observations from the same panoramic video depict the
same physical entity.

USER
Observation A: time={time_a}, direction={direction_a}, description={desc_a}
Observation B: time={time_b}, direction={direction_b}, description={desc_b}
Angular separation: {separation}
Temporal gap: {gap}

Compare the supplied views using distinctive appearance, carried objects,
nearby landmarks, camera motion, and motion feasibility. Interpret direction in
a common scene frame: a scene-fixed object may change camera-relative yaw when
the camera rotates. For a moving entity, check whether the temporal gap and
visible motion support a feasible path.

Shared category or color alone is insufficient in a crowd or a scene with
similar objects. Return "same" when specific visual evidence supports identity
continuity, "different" when the observations conflict, and "unsure" when a
look-alike remains plausible. State the visual cue that determines the verdict.

OUTPUT
{"verdict": "same|different|unsure", "evidence": "..."}
\end{lstlisting}

\subsection{Constructing and Refining Questions}

\subsubsection{Evidence-relation proposal}

\begin{lstlisting}[style=synthesisprompt]
SYSTEM
Propose question structures grounded in a sphere-time evidence memory.

USER
Video duration: {duration}
ERP overview: {overview_frames}
Evidence memory: {evidence_memory}
Question-type guide: {question_type_guide}

Propose {count} question structures supported by the video. Begin from the
evidence rather than from a generic question template. Each proposal must name
the observations that determine its answer and the inference performed over
them. A single observation may support appearance or fine-detail perception.
Multiple observations may support spherical relations, viewpoint changes,
action and state changes, cross-view identity, trajectories, temporal order,
counting, or cross-time aggregation.

Place every proposal on both the sphere and time axes. Record the evidence
times and directions, and state whether the answer comes from one local view or
requires connecting observations across directions or moments. Do not reduce a
dynamic video to a static panorama when its temporal evolution provides the
decisive relation. At the same time, retain strong local questions when fine
detail itself is the challenge.

Use only entities, observations, times, and directions present in the evidence
memory. Vary the entity combinations and inference types instead of producing
paraphrases of the same relation.

OUTPUT
{"proposals": [{"type": "...", "entity_ids": [...],
"observation_ids": [...], "evidence_times": [...],
"evidence_directions": [...], "answer_fact": "...",
"reasoning_requirement": "..."}]}
\end{lstlisting}

\subsubsection{Question and option writing}

\begin{lstlisting}[style=synthesisprompt]
SYSTEM
Write one four-choice question from verified panoramic-video evidence.

USER
Supporting evidence: {supporting_evidence}
Related entities and events: {context_evidence}
Answer fact: {answer_fact}
Reasoning requirement: {reasoning_requirement}

Write a natural question whose answer is fixed by the supporting evidence.
Refer to entities through visible appearance, action, or event context that
identifies them across the complete video. When a description could match
several entities, add a visible temporal or event anchor. Use natural
observer-relative language when direction matters, without exposing numeric
coordinates, timestamps, projection terms, or annotation labels.

Construct three plausible distractors from content that is genuinely present
in the same video. Each distractor should preserve the scene context while
changing one decisive attribute, entity binding, direction, temporal relation,
or event. A viewer who inspects only part of the sphere or timeline should find
these alternatives credible.

Keep all four options parallel in grammar, length, and specificity. Avoid an
option that stands out through extra detail, a matched answer-reversal pair, or
four choices that differ only by one direction word. Exactly one option must be
supported. If the evidence cannot yield a unique question with three credible
distractors, mark the proposal for removal.

OUTPUT
{"question": "...", "options": ["...", "...", "...", "..."],
"gold_index": 0, "distractor_rationales": ["...", "...", "..."],
"drop": false, "drop_reason": ""}
\end{lstlisting}

\subsubsection{Critic and targeted refinement}

\begin{lstlisting}[style=synthesisprompt]
SYSTEM
You are the {critic_role} in a panoramic-video question refinement pipeline.

USER
Question: {question}
Options: {options}
Proposed answer: {gold_option}
Supporting evidence: {supporting_evidence}
Available visual input: {visual_input}

Evaluate the candidate according to your assigned role:
- semantic: the question expresses the selected evidence relation and has one
  supported answer; the annotated question type matches the required
  inference;
- visual grounding: verify entities, attributes, actions, text, event
  boundaries, spherical directions, ERP seam continuity, and cross-view
  identity. Evaluate the intended answer and all three distractors rather than
  checking the answer in isolation;
- blind: read several shuffled versions without visual input and identify cues
  from wording, option length, specificity, grammar, common knowledge, or
  answer-reversal patterns;
- language quality: check evidence sufficiency, distractor plausibility,
  interval validity, answer remapping after option shuffling, clarity, and
  option parallelism.

Global ERP frames reveal competing instances elsewhere in the video. Dense
frames around the evidence interval verify actions, transitions, and temporal
boundaries. Targeted perspective views resolve fine details, spherical
relations, seam cases, and identity matches.

Identify the field responsible for a problem and propose a focused revision
that preserves the verified evidence relation. If the evidence interval or
supporting observations change, request updated views before re-evaluation.

OUTPUT
{"verdict": "pass|revise|drop", "issue": "...", "evidence": "...",
"target_field": "question|options|evidence", "revision": "..."}
\end{lstlisting}

\subsubsection{Joint option audit}

\begin{lstlisting}[style=synthesisprompt]
SYSTEM
Audit a completed panoramic-video question against the supplied video views.

USER
Question: {question}
Options: {options}
Answer: {gold_option}
CoT, if available: {cot}

Evaluate all four options jointly. For each option, identify the observations
that support or contradict it. Confirm that the annotated answer is uniquely
supported, the distractors fail for specific visible reasons, and the CoT uses
the same evidence relation as the QA pair. Request revision when the fact is
sound but wording or option construction is faulty; reject the candidate when
the video supports another answer or more than one option.

OUTPUT
{"verdict": "keep|revise|reject", "reason": "...",
"affected_option": "A|B|C|D|null"}
\end{lstlisting}

\subsection{Exploring the Video}

\subsubsection{Blind exploration}

\begin{lstlisting}[style=synthesisprompt]
SYSTEM
Solve a multiple-choice question by actively inspecting a panoramic video.

USER
Question: {question}
Options: {options}
Video span: {start_time} to {end_time}
ERP overview frames: {overview_frames}

The overview frames sample the complete timeline and preserve all viewing
directions in ERP form. Their left and right boundaries are continuous on the
sphere. They provide global coverage but may not resolve small objects, text,
faces, or fine interactions.

Use panorama(time) to inspect the complete sphere at another moment. Use
look(time, yaw, pitch, fov) to obtain a perspective view of a local region,
either at one moment or across a short interval. Determine what evidence would
distinguish the options before selecting inspections. Search across time and
direction as needed, maintain entity identity between views, and revise the
hypothesis when an observation contradicts it. Finish with one option letter
supported by the inspected video.
\end{lstlisting}

\subsubsection{Evidence-guided exploration}

\begin{lstlisting}[style=synthesisprompt]
SYSTEM
Solve a panoramic-video question using active visual inspection and textual
evidence cues.

USER
Question: {question}
Options: {options}
ERP overview frames: {overview_frames}
Evidence cues: {evidence_descriptions_with_time_yaw_pitch}

Each cue describes a verified observation and gives its time, yaw, and pitch;
it does not provide the answer. Use the cues to prioritize inspections, then
verify the relevant visual content through panorama(time) or
look(time, yaw, pitch, fov). Follow an entity across cues when identity matters
and compare multiple moments when the question asks about order, change, or
aggregation. The final option must follow from the inspected observations, not
from the cue text alone.
\end{lstlisting}

\subsection{Reconstructing and Certifying the CoT}

\subsubsection{Trajectory distillation}

\begin{lstlisting}[style=synthesisprompt]
SYSTEM
Extract the reasoning structure from a successful panoramic-video exploration
trajectory.

USER
Question: {question}
Options: {options}
Answer: {gold_option}
Exploration trajectory: {trajectory}
Executed inspections: {inspection_records}

Separate reasoning from navigation. Repeated bearing adjustments, temporal
resampling, and views that merely reconfirm an established fact belong to the
search process. A decisive inspection changes the current answer, resolves an
ambiguity, or rules out a credible option.

Identify the search goal and the smallest set of decisive inspections. For
each, describe the concrete visual content and the inference it supports. Keep
a revised expectation when the new observation changes the direction of the
solution. Record a detour only when it corresponds to a hypothesis that was
visually tested and rejected. Distinguish evidence that was read directly from
evidence that was partly obscured or inferred from context.

OUTPUT
{"goal": "...", "decisive_observations": [{"inspection_id": 1,
"observation": "...", "implication": "..."}],
"revised_expectation": "...|null",
"useful_detour": "...|null",
"eliminations": [{"option": "A", "reason": "..."}],
"evidence_strength": "direct|partly_obscured|inferred"}
\end{lstlisting}

\subsubsection{CoT writing}

\begin{lstlisting}[style=synthesisprompt]
SYSTEM
Write a concise reasoning trace from a verified exploration trajectory.

USER
Question: {question}
Options: {options}
Answer: {gold_option}
Distilled trajectory: {distilled_trajectory}
Executed inspections: {inspection_records}

Write a continuous first-person account of how the evidence leads to the
answer. Follow the order in which the answer was established:
1. state what evidence would distinguish the options;
2. introduce each decisive moment and viewing direction in natural language;
3. describe the visible content that matters for the decision;
4. compare that content with the requirement of each plausible option;
5. conclude with the answer.

Use concrete visual observations rather than verdict-only phrases. Express
time and direction as a viewer would, not as tool syntax or coordinate tags.
Include a revised expectation only when it explains why later evidence changed
the conclusion. Omit repeated inspections and navigation that did not affect
the reasoning. Match confidence in an observation to its visual clarity, while
keeping the final conclusion decisive when the option comparison is complete.

OUTPUT
{"cot": "..."}
\end{lstlisting}

\subsubsection{Visual passage attribution}

\begin{lstlisting}[style=synthesisprompt]
SYSTEM
Link every visual statement in a CoT to the inspection that supports it.

USER
CoT: {cot}
Executed inspections: {inspection_records}

For each inspection, quote the exact passage that describes what was observed
there. Match the passage by its stated time, direction, and visual content.
Return null when an inspection is not used in the CoT. Do not assign one
passage to several inspections unless it explicitly combines their evidence.

OUTPUT
{"matches": [{"inspection_id": 1, "passage": "...|null"}]}
\end{lstlisting}

\subsubsection{CoT certification}

\begin{lstlisting}[style=synthesisprompt]
SYSTEM
Certify whether a panoramic-video CoT is sufficient and faithful to its visual
trajectory.

USER
Question: {question}
Options: {options}
Answer: {gold_option}
CoT without its final answer line: {cot_body}
Attributed inspections: {attributed_inspections}

Perform two checks:
1. Sufficiency: infer the answer from the CoT body and report whether it
   recovers the annotated option. Judge the reasoning rather than the presence
   of an answer phrase.
2. Time-direction faithfulness: compare each visual passage with the timestamp,
   yaw, pitch, and field of view of its attributed inspection. Flag relocated
   evidence, reversed directions, or temporal drift.
OUTPUT
{"sufficient": true, "recovered_option": "A",
"faithful": true, "mismatches": [], "reason": "..."}
\end{lstlisting}

\end{document}